\documentclass[11pt]{article}
\usepackage[letterpaper,margin=1in]{geometry}
\usepackage[utf8]{inputenc}
\usepackage[T1]{fontenc}
\usepackage{textcomp}
\usepackage{mathptmx}
\usepackage{amsmath,amssymb}
\usepackage{newunicodechar}
\newunicodechar{−}{\ensuremath{-}}
\newunicodechar{≈}{\ensuremath{\approx}}
\newunicodechar{Δ}{\ensuremath{\Delta}}
\newunicodechar{α}{\ensuremath{\alpha}}
\newunicodechar{γ}{\ensuremath{\gamma}}
\newunicodechar{η}{\ensuremath{\eta}}
\newunicodechar{σ}{\ensuremath{\sigma}}
\newunicodechar{→}{\ensuremath{\rightarrow}}
\usepackage{graphicx}
\usepackage{longtable,booktabs,array,calc}
\usepackage{caption}
\usepackage{setspace}
\usepackage{xurl}
\usepackage[hidelinks]{hyperref}
\usepackage{etoolbox}

\makeatletter
\def\maxwidth{\ifdim\Gin@nat@width>\linewidth\linewidth\else\Gin@nat@width\fi}
\makeatother
\setkeys{Gin}{width=\maxwidth,height=\textheight,keepaspectratio}
\begin{document}
\begin{center}
{\Large\bfseries Smartphone-Enabled Depth-Aware Transillumination Imaging for Detection of Chicken Breast Fillet Myopathies in Chicken Fillets\par}
\vspace{10pt}
{\normalsize Chaitanya Pallerla$^{a,b}$, Ramesh Bahadur Bist$^{a}$, Yihong Feng$^{a}$, Yang Tian$^{a}$, Shiva Paudel$^{a}$, Casey M. Owens$^{c}$, Cengiz Koparan$^{a}$, Jeyamkondan Subbiah$^{b}$, Dongyi Wang$^{a,b,*}$\par}
\vspace{6pt}
{\small
$^{a}$ Department of Biological and Agricultural Engineering, University of Arkansas, Fayetteville, AR, USA, 72701\par
$^{b}$ Department of Food Science, University of Arkansas, Fayetteville, AR, USA, 72701\par
$^{c}$ Department of Poultry Science, University of Arkansas, Fayetteville, AR, USA, 72701\par
$^{*}$ Corresponding: \href{mailto:dongyiw@uark.edu}{dongyiw@uark.edu}\par}
\end{center}
\textbf{Abstract}

Woody Breast (WB) myopathy is a major quality defect in broiler chicken characterized by fibrosis, increased tissue hardness, and altered structural properties that negatively affect product quality and consumer acceptance. Conventional palpation is subjective and labor-intensive, while many non-destructive approaches require specialized imaging equipment or substantial computational resources. This study developed a portable smartphone-based transillumination and 3D point-cloud framework with multimodal machine learning for non-destructive WB severity classification. A total of 267 chicken breast fillets representing normal breast (NB, \emph{n} = 87), mild WB (MWB, \emph{n} = 85), and severe WB (SWB, \emph{n} = 95) were evaluated using RAW transillumination images and full-resolution 3D point clouds acquired with an iPhone 16 Pro. A total of 91 features, including 14 image-derived, 76 geometric, and one physics-informed Beer-Lambert attenuation feature, were extracted. Partial correlation analysis controlling for thickness demonstrated a significant association between image-derived characteristics and WB severity independent of fillet geometry (\emph{r} = 0.413, \emph{p} \textless{} 0.0001), supporting the presence of optical information beyond dimensional differences. Thickness-normalized analysis further showed that SWB tissue exhibited approximately 26\% greater light attenuation per unit thickness than NB, indicating that transillumination differences were not solely attributable to fillet thickness. For classification models, the 2D-only models achieved approximately 82\% test accuracy, whereas the final multimodal ensemble combining Fusion-DGCNN, XGBoost, and physics-guided logistic regression (PGLR) achieved approximately 90\% accuracy. The findings demonstrate the complementary value of optical, 3D geometric, and physics-informed information and support the feasibility of a consumer-grade smartphone as a portable, non-destructive platform for WB severity assessment.

\textbf{Keywords:} Woody breast myopathy, Transillumination imaging, Multimodal machine learning, Point cloud analysis, Mobile Computing

\hypertarget{introduction}{%
\section{Introduction}\label{introduction}}

Chicken meat is one of the most consumed animal protein sources globally due to its affordability, availability, and favorable nutritional profile, including high protein content and low-fat levels (Petracci et al., 2019). In response to increasing consumer demand, the poultry industry has aggressively selected broilers for rapid growth and high breast meat yield. However, this genetic intensification has been accompanied by a sharp rise in growth-related muscular myopathies such as woody breast (WB), white striping (WS), and spaghetti meat (SM) (Soglia et al., 2016; Petracci et al., 2019; Caldas-Cueva \& Owens, 2020).

WB is characterized by increased tissue hardness, fibrosis, and collagen infiltration, leading to downgraded meat quality and functional impairments in texture and water-holding capacity (Sihvo et al., 2014; Mudalal et al., 2015). SM defect displays separation of muscle fibers resembling cooked pasta strands and is associated with poor muscle integrity and decreased protein content (Baldi et al., 2018). The co-occurrence and overlapping incidence of WB and SM complicate diagnosis and quality control (Soglia et al., 2019).

Traditional assessment of these defects relies heavily on subjective tactile palpation by trained personnel, which is time-consuming, labor-intensive, and often inconsistent (Caldas-Cueva et al., 2021). This has prompted the development of objective, non-destructive, and automatable detection methods. Recent studies have explored technologies such as near-infrared spectroscopy (Wold et al., 2017), hyperspectral imaging (Pallerla et al., 2024), image coherence tomography (Yoon et al., 2016), and light scattering imaging (Cai \& Lu, 2024) for WB detection. These methods leverage changes in image or structural tissue properties arising from pathological remodeling. However, despite their effectiveness, these techniques often face practical challenges related to high equipment cost, complex calibration requirements, limited portability, and constrained suitability for real-time, high-throughput deployment in industrial poultry processing environments.

To highlight, light scattering imaging (LSI) has emerged as a promising technique for differentiating between normal and WB-affected fillets. Cai and Lu (2024) demonstrated that scattering features derived from broadband light interactions could yield classification accuracies exceeding 90\% when optimized through feature selection and regularized discriminant analysis. This reflects the altered tissue morphology and heterogeneity associated with WB, which modulates diffuse reflectance behavior. Caldas-Cueva et al. (2021) applied image analysis techniques to identify WB severity in broiler carcasses, achieving over 90\% classification accuracy. These approaches show the value of spatial and morphological imaging features in automating WB defect detection. However, the challenge remains in integrating these techniques into in-line processing environments while ensuring robustness to sample variability.

Transillumination imaging, a straightforward and powerful imaging modality, offers a potentially cost-effective solution for rapid defect detection in chicken meat. In this technique, light is projected through the sample, and a camera captures the transmitted light pattern on the opposite side. As light passes through biological tissues, it undergoes absorption, scattering, and diffusion, which depend on the physical and biochemical properties of the tissue matrix. Defects such as WB, characterized by increased connective tissue and reduced transparency, are likely to scatter and attenuate transmitted light more than normal fillets. Recent studies support this premise by detailing how altered muscle composition and structure influence image properties. Wold et al. (2017) demonstrated that WB fillets absorb and scatter NIR light differently than normal tissues, reflecting biochemical changes due to fibrosis. Yoon et al. (2016) further revealed through OCT that structural disruptions in muscle tissue modulate subsurface light reflection patterns. These findings reinforce that tissue heterogeneity, and refractive index mismatches play a significant role in image signal variation and confirm that myopathies significantly affect light propagation.

Understanding material-light interactions is critical for optimizing the transillumination system. The absorption of visible or near-infrared light by muscle proteins, lipids, and interstitial water, combined with multiple scattering events caused by tissue inhomogeneities, governs the detected image. For instance, WB tissue, rich in fibrotic and collagenous material, has different refractive index mismatches compared to healthy muscle, which influences light path length and scattering angles. By analyzing transillumination patterns, several quantitative features can be extracted: mean transmission intensity, spatial variance, gradient maps, and morphological indicators such as area and perimeter of dense/loose regions. WB fillets are expected to exhibit lower transmission intensities and higher uniformity due to dense tissue composition. The principal novelty of this study lies in the integration of transillumination imaging and TrueDepth based 3D sensing within a consumer grade smartphone platform for WB severity assessment. To the best of our knowledge, this is the first study to investigate transillumination imaging specifically for multiclass WB severity classification and the first to combine transmitted light characteristics with 3D point cloud derived geometric information for this purpose. Unlike conventional approaches relying on dedicated hyperspectral, spectroscopic, or laboratory imaging equipment, the proposed framework uses an iPhone 16 Pro to acquire high resolution RAW transillumination images and TrueDepth derived 3D information within an integrated acquisition framework. This enables complementary characterization of internal optical variations through 2D transmitted light patterns and external morphology through 3D geometric descriptors. The combination of these two modalities also allows the influence of fillet geometry, particularly tissue thickness, to be distinguished from transillumination characteristics associated with internal tissue alterations. From an engineering perspective, implementation using a widely available consumer device reduces equipment requirements and provides a portable and scalable framework with potential for practical deployment in poultry processing environments.

This study was conducted with three primary objectives to explore the feasibility of using a consumer grade smartphone based multimodal imaging framework for the non-destructive detection and classification of woody breast (WB) severity. The first objective was to develop a low-cost smartphone based transillumination imaging system capable of acquiring both RAW transillumination images and TrueDepth point cloud data for non-destructive assessment of chicken breast fillets. The second objective was to investigate whether transillumination derived optical features can characterize woody breast related fibrosis independently of sample geometry by separating the effects of tissue thickness from internal structural changes. The third objective was to develop and evaluate multimodal machine learning models that integrate transillumination image features with 3D geometric information to accurately classify woody breast severity and determine the contribution of multimodal data fusion toward improving classification performance.

\hypertarget{materials-and-methods}{%
\section{Materials and Methods}\label{materials-and-methods}}

\hypertarget{sample-preparation}{%
\subsection{Sample Preparation}\label{sample-preparation}}

This study was conducted at the University of Arkansas\textquotesingle s Biological and Agricultural Engineering laboratory. Broilers of 8 weeks of age were randomly selected from the University\textquotesingle s poultry farm and processed in the University of Arkansas pilot processing plant. Broilers were processed using commercial methods, and fillets were deboned from carcasses approximately 30 min postmortem. Then, a total of 267 unchilled chicken breast fillets, including the left and right sides, were collected and scored in the processing plant. The samples were classified as normal breast (NB), mild woody breast (MWB), and severe woody breast (SWB) using hand palpation (Tijare et al. 2016) by two experts (NB = 87; MWB = 85; SWB = 95). After WB scoring, the samples were transported to the laboratory for transillumination imaging and 3D point cloud data collection. The processing and WB scoring protocol followed the procedure described by Caldas-Cueva et al. (2021).

\hypertarget{transillumination-imaging-setup}{%
\subsection{Transillumination Imaging Setup}\label{transillumination-imaging-setup}}

\hypertarget{hardware-configuration-and-imaging-device}{%
\subsubsection{Hardware Configuration and Imaging Device}\label{hardware-configuration-and-imaging-device}}

The system utilized a pair of Lightdot UFO LED high-bay lights (fig. 1), each delivering 75,000 lumens at a 5000 K color temperature, providing consistent, bright illumination for imaging.

\includegraphics[width=3.94444in,height=2.47511in]{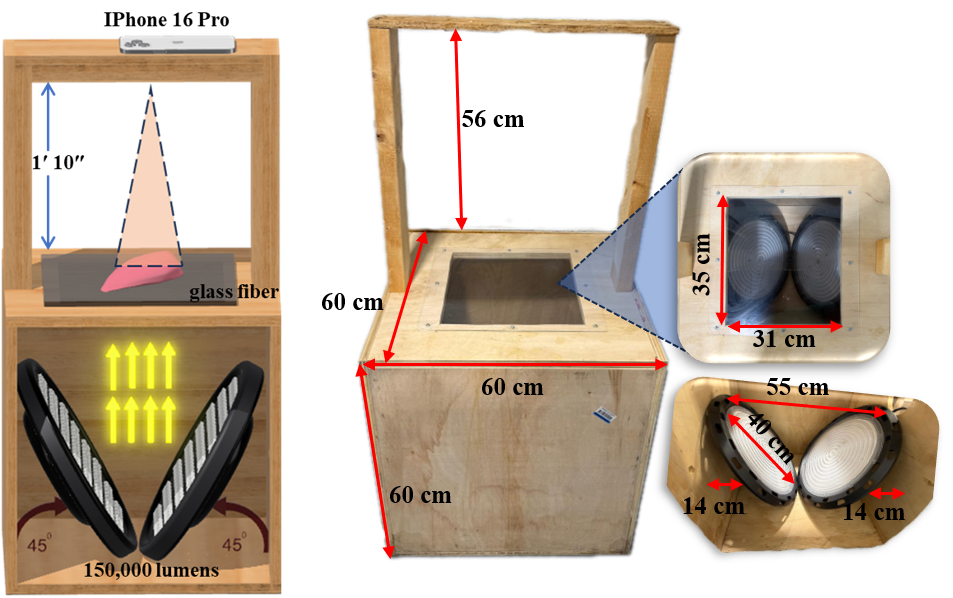}

Fig. 1 Transillumination imaging setup for WB detection

The lights were placed below a transparent matte glass fiber diffusion sheet to ensure uniform light scattering across the imaging area (35 x 31 cm) and the light source is 55 cm from the point of convergence of two lights to the glass fiber. Two high-output LED panels, each providing 75,000 lumens, were positioned at 45° so that their beams converged onto a single focal region on the diffuser surface. This arrangement delivered a combined luminous flux of approximately 150,000 lumens at the image convergence point. The chicken fillet sample was placed precisely at this region, where the overlapping beams enhanced forward-scattered transmission through the tissue. By concentrating total illumination in this manner, the configuration produced a pseudo-coherent lighting effect that increased image contrast and improved sensitivity to fine internal structural variations such as collagen deposition and muscle fiber separation that are characteristic of WB myopathy.

Imaging acquisition was conducted using an Apple iPhone 16 Pro equipped with a 24 mm focal length Fusion Camera, a wide f 1.78 aperture for RAW color image acquisition and TrueDepth sensor for point cloud data acquisition. The camera was mounted vertically using a fixed rig at a standardized height of 1 ft 10 in to maintain consistent imaging geometry across all samples. All captures were conducted inside a light-sealed wooden enclosure to block ambient light and maximize the reliability of transmitted signal detection. Fig. 2 shows the transilluminated images of NB, MWB, SWB with internal structure and changes in transmitted light from each category.

\includegraphics[width=4.25018in,height=2.10534in]{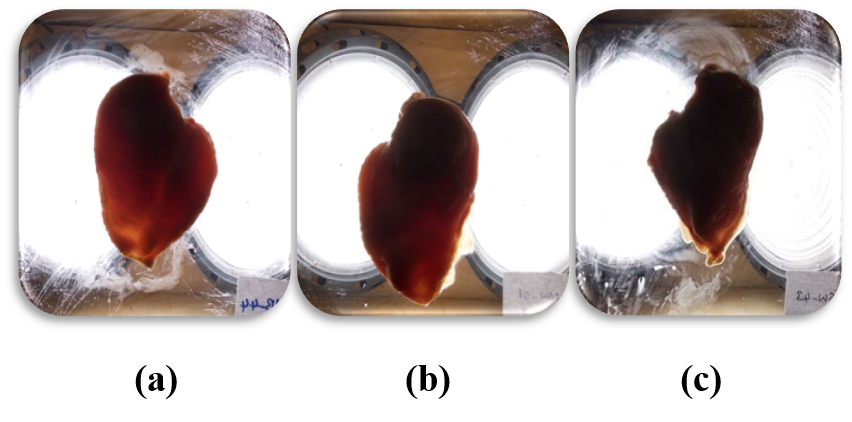}

Fig. 2 Transilluminated images of chicken breast fillet with different WB severity levels. (a) Normal Breast (NB); (b) Mild Woody Breast (MWB); (c) Severe Woody Breast (SWB)

\hypertarget{raw-image-acquisition-and-preprocessing-before-feature-extraction}{%
\subsubsection{RAW Image Acquisition and Preprocessing Before Feature Extraction}\label{raw-image-acquisition-and-preprocessing-before-feature-extraction}}

All images were acquired using the Apple RAW DNG format to retain full sensor data and maximum dynamic range of approximately 14 bits using the Halide Mark II App (Version 2.17.5; Lux Optics Incorporated, USA). The RAW format preserves linear pixel intensity values and avoids in-camera enhancements such as sharpening, tone mapping, or contrast adjustments, making it ideal for capturing subtle grayscale variations resulting from light tissue interactions. Typical imaging parameters included a resolution of up to 5424 x 4068 pixels, fixed ISO 50, and exposure times ranging from 1/14,000 seconds to enable high speed and blur free imaging under intense transillumination conditions.

The RAW image will be preprocessed, and then for handcrafted image feature extraction which will be discussed in Section 2.3.2.1. Figure 3 Output 1 shows the preprocessing pipeline, which aims to remove background elements such as light reflections and chamber edges that can introduce noise in downstream analysis.

\includegraphics[width=6.47639in,height=2.36845in]{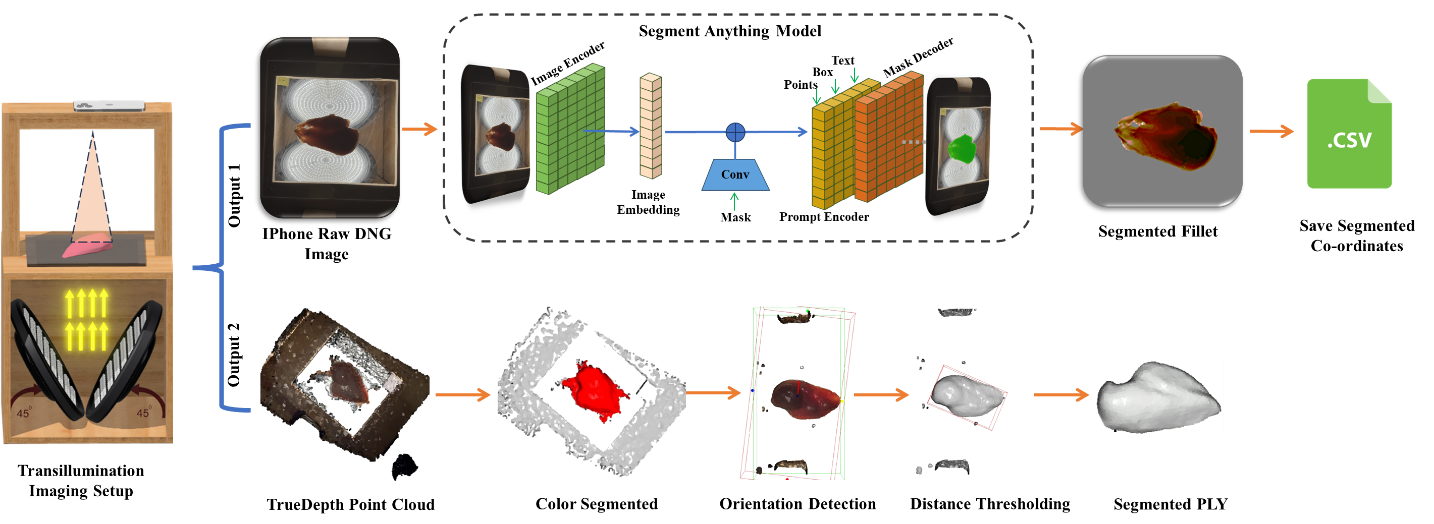}

Fig. 3 Data preprocessing workflow prior to feature extraction

Following image acquisition, the chicken breast fillet was automatically segmented from the background using the Segment Anything Model (SAM) with the Vision Transformer Base (ViT-B) backbone (Kirillov et al., 2023). A single positive point prompt was provided at the geometric center of each image, corresponding to the approximate center of the fillet, which was consistently positioned at the center of the imaging chamber during image acquisition. No negative point prompts, bounding box prompts, or text prompts were used. The SAM predictor generated three candidate segmentation masks , each associated with a confidence score. The mask with the highest confidence score was automatically selected as the final segmentation result for subsequent analysis. The selected binary mask was then applied to isolate the fillet from the background while preserving the original pixel intensities and spatial coordinates of all foreground pixels. Background pixels outside the segmented region were removed and replaced with a transparent or zero-valued background, ensuring that subsequent preprocessing and handcrafted feature extraction were performed exclusively on the segmented fillet region without interference from background artifacts, reflections, or imaging chamber components. All segmentation was performed using the publicly available SAM ViT-B model (sam\_vit\_b\_01ec64.pth) accelerated with NVIDIA CUDA enabled GPU computation to ensure efficient processing.

\hypertarget{point-cloud-acquisition-and-auto-segmentation}{%
\subsubsection{Point Cloud Acquisition and Auto Segmentation}\label{point-cloud-acquisition-and-auto-segmentation}}

Three-dimensional point clouds were acquired using the 3D Scanner App (Version 2.3; Laan Labs, New York, NY, USA) running on an Apple iPhone 16 Pro. Because the chicken breast fillets were relatively small objects, the application was operated in TrueDepth mode, which utilizes the iPhone\textquotesingle s front facing TrueDepth structured light depth sensing system rather than the rear LiDAR scanner. Each fillet was scanned for approximately 10 s while the device slowly moved around the sample to ensure complete surface coverage. The application was operated using its default TrueDepth acquisition settings, and the resulting point clouds were exported directly for subsequent processing. No additional app-based filtering, mesh refinement, smoothing, or post processing operations were applied within the 3D Scanner App prior to export. The exported point cloud data were subsequently processed using a custom five-stage 3D point cloud preprocessing pipeline as illustrated in Fig.3 .

The first stage shows the raw 3D scanner output containing the chicken breast fillet along with the surrounding scanning environment including the platform, frame structures, and background clutter captured by the depth sensor. The second stage demonstrates semantic segmentation, where algorithms isolate the fillet from environmental noise using techniques including RANSAC plane detection (Fischler et al., 1981) to remove the platform, Euclidean clustering to group connected points, and statistical outlier removal to eliminate sensor artifacts.The third stage produces an isolated breast fillet represented as a clean 3D point cloud enclosed within a bounding box. In this stage, the principal orientation of the fillet is estimated by analyzing the spatial distribution of the point cloud, allowing the determination of both its current orientation and the desired canonical orientation. This orientation information provides a reliable estimate of the fillet direction, which is essential for consistent downstream geometric analysis. The fourth stage uses the orientation parameters obtained in the previous stage to perform distance-based margin cropping along the x, y, and z axes. Cropping is guided by thresholding (5cm) relative to the dominant point cluster corresponding to the fillet body. The 5 cm margin cropping threshold was selected empirically based on repeated visual inspection of the scanned point clouds obtained from the experimental dataset. Following orientation normalization, the majority of residual points outside the main fillet corresponded to fragments of the supporting surface, scanning artifacts, or isolated noise generated near the specimen boundaries. A threshold of 5 cm relative to the dominant point cluster consistently preserved the complete chicken breast fillet while effectively removing these unwanted peripheral points across all samples. The resulting output is a fully isolated and standardized fillet point cloud that will be used for the extraction of 76 handcrafted 3D geometric features, which will be described in the Section 2.3.2.2.

\hypertarget{model-architecture-for-multimodality-feature-extraction-and-wb-severity-classification}{%
\subsection{Model Architecture for Multimodality Feature Extraction and WB Severity Classification~}\label{model-architecture-for-multimodality-feature-extraction-and-wb-severity-classification}}

\hypertarget{model-summary}{%
\subsubsection{Model Summary}\label{model-summary}}

This work employs a hybrid modeling framework that combines deep geometric learning, gradient-boosted decision trees, and a physics-guided linear model. A customized Fusion- Dynamic Graph Convolutional Neural Network (DGCNN) model built on a dynamic graph convolutional architecture is used to process 2D backlighting features using Multi-Layer Perceptron (MLP) together with full-resolution 3D point clouds, utilizing all points from each PLY file without sampling and dynamically rebuilding its k-nearest-neighbor graph in feature space at every layer to learn a global geometric representation. In parallel, XGBoost is trained on the full scaled handcrafted feature set, and a Physics-Guided Logistic Regression (PGLR) is trained on a small, dedicated subset of 5 physics-motivated features dense-area percentage, attenuation-coefficient, and 3 of local-thickness heterogeneity statistics (PC\_Local\_Thickness\_Mean, PC\_Local\_Thickness\_Std, PC\_Local\_Thickness\_CV) kept deliberately simple for interpretability. The final prediction is obtained through a soft voting ensemble that combines the class probability outputs of Fusion-DGCNN, XGBoost, and PGLR, combining the complementary strengths of deep geometric learning, feature-based gradient boosting, and physically-grounded linear modeling for improved classification performance, as shown in Fig. 4.

As comparison, four additional point-cloud architectures Fusion-PointNet, Pure PointNet (3D-only, no 2D fusion), PointNet++, and a PointBERT-style patch-transformer model were trained under the same full-resolution, no-subsampling protocol as Fusion-DGCNN. Classical machine learning models LightGBM and a standalone MLP were also trained independently on the same scaled combined feature set. Every possible ensemble combination across all nine trained models (Fusion-PointNet, Pure PointNet, PointNet++, Fusion-DGCNN, PointBERT-style, XGBoost, LightGBM, MLP, and PGLR) was evaluated by nested cross-validation out-of-fold accuracy.

\includegraphics[width=6.5in,height=3.28107in]{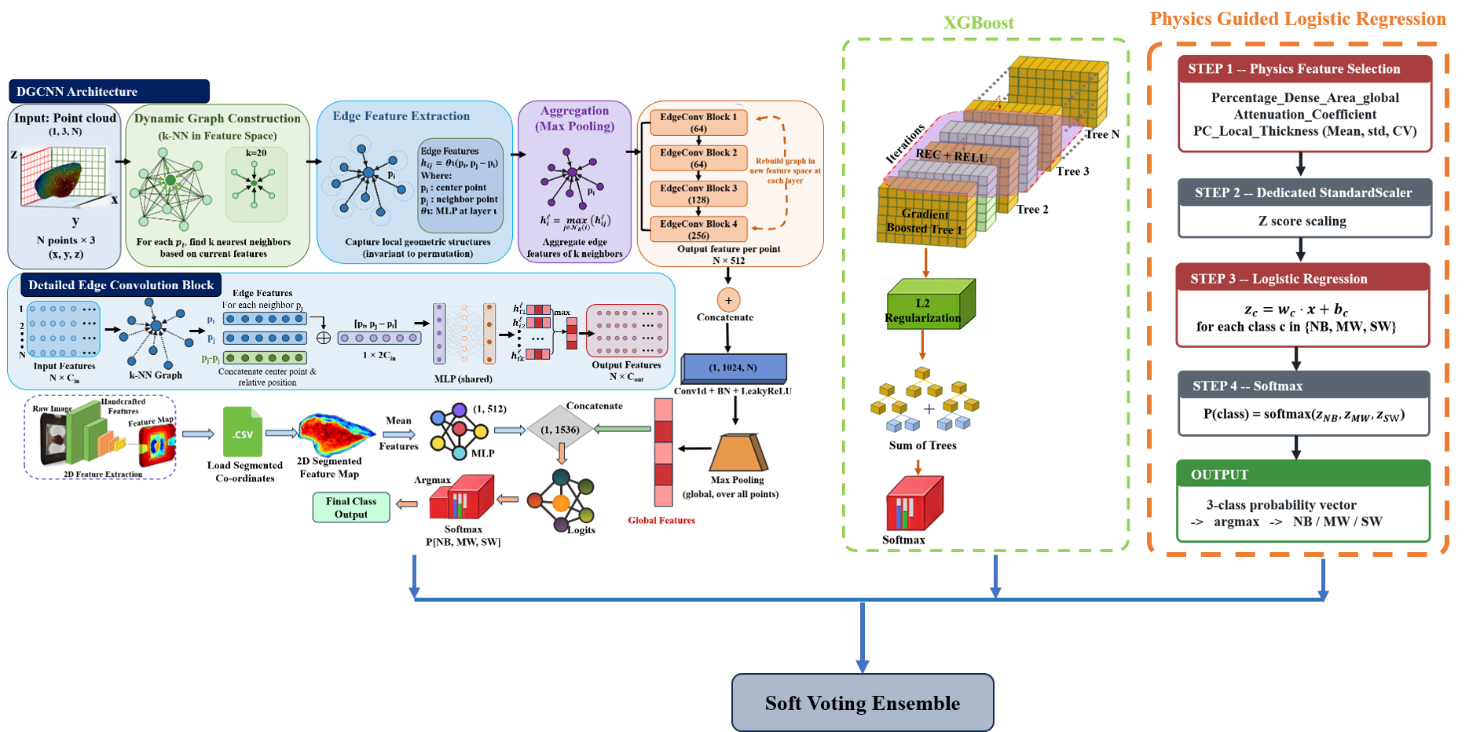}

Fig. 4 The proposed Fusion DGCNN, XGBoost and PGLR Soft Voting Ensemble Architecture for WB Severity Classification

\hypertarget{traditional-machine-learning-model}{%
\subsubsection{Traditional Machine Learning Model}\label{traditional-machine-learning-model}}

\hypertarget{handcrafted-raw-rgb-features-for-traditional-machine-learning-models}{%
\paragraph{Handcrafted RAW RGB features for Traditional Machine Learning Models}\label{handcrafted-raw-rgb-features-for-traditional-machine-learning-models}}

Once RAW preprocessing was completed, a comprehensive set of image-derived features was extracted exclusively from the segmented fillet region. These features included spatial texture descriptors such as Gabor filter responses computed at multiple orientations, gradient-based metrics including mean and standard deviation of edge strength, and texture roughness quantified using local variance measures. Edge focused features were also extracted, including total edge pixel counts, percentage of strong edges, gradient magnitude histogram bin distributions, and dominant edge orientation angle expressed in degrees. In addition to quantitative features, each sample generated a pseudo-colored visualization of gradient magnitude to support qualitative assessment and visual inspection. All extracted quantitative features were stored in structured CSV files to facilitate downstream statistical analysis and machine learning model training.

The extracted features are as follows:

\textbf{Percentage\_Dense\_Area} measures the proportion of dark, highly attenuating pixels. This metric is expected to be higher in WB due to collagen-dense areas that reduce light transmission. It is calculated using Eq. (1).

\begin{quote}
\(\text{Percentage}\text{\_}\text{Dense}\text{\_}\text{Area} = \frac{N_{\text{dark pixels}}}{N_{\text{total fillet pixels}}} \times 100\) (1)
\end{quote}

\(N_{dark}\) denotes the number of fillet pixels classified as low intensity pixels.

\(N_{\text{total fillet pixels}}\) denotes the total number of pixels belonging to the chicken breast fillet.

To determine this, grayscale intensity values are first extracted from the RGB image and normalized only within the fillet region using min-max normalization to 0\textasciitilde255. A fillet pixel \(p\) is classified as dark if its normalized grayscale intensity \(I_{p}^{norm}\) satisfies

\[I_{p}^{norm} < T_{dark}\]

where \(T_{dark}\) is a fixed intensity threshold set to a value of 10. This threshold isolates regions with minimal transmitted light, corresponding to image-dense tissue areas. The dark pixel mask is further refined using morphological closing to remove small gaps and ensure spatial continuity of dense regions.

\textbf{Attenuation Coefficient} is a physics-based feature that approximates the tissue\textquotesingle s linear attenuation coefficient by normalizing the dense-area signal (Eq. (1)) by fillet thickness, analogous to the relationship between light attenuation and optical path length described by the Beer-Lambert law. The classical Beer-Lambert law strictly holds only in non-scattering, purely absorbing media; biological tissue is highly scattering, and the law does not hold in its unmodified form under these conditions (Oshina \& Spigulis, 2021). Accordingly, this feature is presented as an attenuation-coefficient analogue rather than a direct application of the classical law. It preserves the law\textquotesingle s core physical principle that an attenuation signal must be normalized by the optical path length it accumulated over to yield an intrinsic, thickness-independent tissue property without invoking its exponential/logarithmic form. This normalization principle is retained even in models developed specifically for thick, scattering biological tissue, which likewise define their derived attenuation parameters relative to path length while introducing additional terms to account for scattering (Bhatt et al., 2016). `Percentage\_Dense\_Area' alone conflates the amount of attenuating tissue with the fillet\textquotesingle s physical thickness. Two fillets can show the same dense-area percentage despite very different thicknesses and are not equivalently attenuating per unit path length in that case. Rescaling dense-area by thickness yields a thickness-independent proxy for tissue optical density. The metric, calculated in Eq. (2), is expected to be higher in Woody Breast fillets due to their combination of denser, more collagenous tissue and increased overall thickness.

\(\text{Attenuation~Coefficient}\text{ = }\frac{\text{Percentage~Dense~Area}}{\text{Thickness}}\) (2)

where \(\text{Thickness}(mm)\) represents the Z-axis extent of the fillet and is calculated from the three-dimensional point cloud as\(\ \text{Thickness}\text{ = }Z_{\max} - Z_{\min}\) where \(Z_{\max}\) and \(Z_{\min}\) are the maximum and minimum Z-coordinates, respectively, of the fillet point cloud.

\textbf{Gabor\_Mean\_0, 45, 90, 135} are average intensity responses from Gabor filters at four key orientations (Jain and Farrokhnia, 1991; Manjunath and Ma, 1996). In this study, Gabor features are extracted by applying a 2D complex Gabor filter to the normalized grayscale image \(I(x,y)\). The filter is applied at a fixed spatial frequency \(f = 0.2\) and at four orientations \(\theta \in \{ 0,\pi/4,\pi/2,3\pi/4\}\). For each orientation, the real and imaginary filter responses \(R_{\theta}(x,y)\) and \(I_{\theta}(x,y)\) are obtained using convolution. These features detect directional muscle fiber textures, disrupted in WB. The phase-independent Gabor response is computed as Eq. (3).

\(G_{\theta}(x,y) = \sqrt{R_{\theta}(x,y)^{2} + I_{\theta}(x,y)^{2}}\) (3)

The final Gabor feature for each orientation is defined as the spatial mean of the response over the fillet region and calculated as Eq. (4).

\({Gabor\_ Mean}_{\theta} = \frac{1}{N_{fillet}}\sum_{i \in \Omega_{fillet}}^{}G_{\theta}(i)\) (4)

Where I(x,y) is the normalized grayscale image derived from the DNG file. \(f\)is the spatial frequency of the Gabor filter and controls the scale of texture being detected. \(\theta\) is the orientation of the filter and determines directional sensitivity. \(\ R_{\theta}(x,y)\)and \(I_{\theta}(x,y)\) are the real and imaginary filter responses. \(\ G_{\theta}(x,y)\)is the magnitude of the complex Gabor response and represents orientation specific texture energy. \(\Omega_{fillet}\)is the set of pixels belonging to the segmented fillet. \(N_{fillet}\)is the total number of fillet pixels.

\textbf{Mean\_Gradient\_Magnitude} (\(\mu\)) and \textbf{Std\_Gradient\_Magnitude} (\(\sigma\)) measure the overall gradient strength and variability (Xue et al., 2014; Canny et al., 1986), reflecting the presence of sharp structural transitions in fibrotic WB tissue. The horizontal and vertical image gradients are computed using the Sobel operator applied to the grayscale image \(I(x,y)\). It is calculated as seen in Eq. (5).

\(G_{x} = \frac{\partial I}{\partial x},G_{y} = \frac{\partial I}{\partial y}\) (5)

Here,\(\ G_{x}\) represents the intensity change along the horizontal direction, and\(\ G_{y}\) represents the intensity change along the vertical direction. The gradient magnitude at pixel \(i\ \)is defined as seen in Eq. (6).

\begin{quote}
\(\text{Magnitude} = \sqrt{G_{x}^{2} + G_{y}^{2}}\) (6)
\end{quote}

The mean gradient magnitude is defined as seen in Eq. (7), where \(N\ \)denotes the total number of pixels within the segmented fillet region. The index \(i\) refers to the \(i^{th}\)pixel within this region.

\(\mu = \frac{1}{N}\sum_{i = 1}^{N}{Magnitude}_{i}\) (7)

The standard deviation of gradient magnitude is defined as seen in Eq. (8).

\begin{quote}
\(\sigma = \sqrt{\frac{1}{N}\sum_{i = 1}^{N}\left( {Magnitude}_{i}-\mu \right)^{2}}\) (8)
\end{quote}

These metrics quantify the overall strength and variability of intensity transitions. Higher values indicate sharper structural changes typical of fibrotic WB tissue, whereas lower values indicate smoother gradients associated with softer muscle tissue.

\textbf{Mean\_Local\_Variance} evaluates local intensity fluctuations. Lower values suggest uniform tissue (WB). Local variance (Haralick et al., 1973) is computed within a sliding window \(W_{i}\) centered at pixel \(i\). The variance within each window is defined as seen in Eq. (9).

\(Var(W_{i}) = \frac{1}{\mid W_{i} \mid}\sum_{j \in W_{i}}^{}\left( I_{j}-\mu_{W_{i}} \right)^{2}\) (9)

where \(\mid W_{i} \mid \ \)is the number of pixels in the window and \(\mu_{W_{i}}\)is the mean intensity within that window. The mean local variance over the fillet region is then defined as seen in Eq. (10).

\begin{quote}
\(Mean\_ Local\_ Variance = \frac{1}{N}\sum_{i = 1}^{N}{Var}(W_{i})\) (10)
\end{quote}

\textbf{Edge\_Pixel\_Count} represents the number of pixels classified as edges using the Canny edge detection algorithm (Canny et al., 1986), which identifies pixels with strong intensity gradients after non-maximum suppression and hysteresis thresholding value of 20.

\textbf{Gradient\_Hist\_Bin\_1 to Bin\_5} describe the distribution of gradient magnitudes (Dalal et al., 2005; Hadjidemetriou et al., 2004) within the fillet region by partitioning the gradient magnitude range into five non overlapping intervals. The count for bin \(k\)is defined as seen in Eq. (11).

\begin{quote}
\({Bin}_{k} = Count\left( g_{i}\in R_{k} \right)\) (11)
\end{quote}

Where \(g_{i}\) is the gradient magnitude at pixel \(i\), and \(R_{k}\) denotes the value range corresponding to bin \(k\). A higher concentration of counts in lower bins indicates smoother tissue with weak intensity transitions, while a broader distribution across higher bins reflects increased heterogeneity and pronounced structural variation associated with WB myopathy.

All the features described above were chosen based on their ability to capture multiscale and multimodal differences in tissue architecture resulting from WB abnormalities.

\hypertarget{handcrafted-point-cloud-for-traditional-machine-learning-models}{%
\paragraph{Handcrafted Point Cloud for Traditional Machine Learning Models}\label{handcrafted-point-cloud-for-traditional-machine-learning-models}}

The 3D point cloud features were extracted from cleaned and centered fillet point clouds stored in PLY format using a deterministic geometric processing pipeline implemented in Python, and these features were exclusively handcrafted and used only for the classical machine learning classifiers. For each sample, the point cloud processing starts by centering the point cloud by subtracting the centroid and normalizing by the maximum radial distance to ensure scale consistency across samples. A total of 18 basic statistical features were computed from the x, y, and z coordinates, including measures of central tendency, dispersion, extrema, and range, to describe overall fillet size and thickness distribution. In addition, thirteen (13) distance-based features were derived from the Euclidean distances of points to the centroid, summarized using mean, variability, extrema, and percentile statistics to capture radial dispersion and geometric irregularity. Covariance analysis of the point coordinates produced 15 covariance and eigenvalue-based features through eigen decomposition, characterizing dominant geometric directions and structural anisotropy using descriptors such as linearity, planarity, sphericity, omnivariance, and eigenentropy. Global shape properties were further quantified using 10 shape features computed from axis-aligned bounding box dimensions and derived metrics such as surface area, volume, compactness, and sphericity-related indices. Finally, seventeen (17) discriminative features were explicitly engineered to encode thickness-normalized geometry by combining bounding box measurements, radial distance statistics, point density, and vertical thickness variability, enabling differentiation between true pathological swelling and normal geometric variation, and 3 local thickness features. All handcrafted point cloud features were concatenated into fixed-length vectors and supplied to traditional classifiers, including XGBoost, LightGBM, MLP, and all fusion-point models.

\hypertarget{types-of-traditional-machine-learning-and-deep-point-cloud-models}{%
\paragraph{Types of Traditional Machine Learning and Deep Point Cloud Models}\label{types-of-traditional-machine-learning-and-deep-point-cloud-models}}

Three machine learning models, namely multilayer perceptron (MLP), XGBoost, and LightGBM, were evaluated using the handcrafted feature set, which consisted of a total 91-dimensional vector combining 2D transillumination features and 3D geometric descriptors extracted from the point clouds. XGBoost constructs an additive ensemble of decision trees, in which each successive tree is trained to correct the residual errors of the preceding ensemble using a second-order approximation of the loss function together with an explicit regularization term controlling tree complexity (Chen et al., 2016). LightGBM similarly builds an additive ensemble of decision trees, but grows each tree leaf-wise, splitting the leaf that yields the greatest reduction in loss rather than growing level-by-level, allowing it to reach lower loss with fewer splits in high-dimensional feature spaces (Ke et al., 2017). Both were tuned using a sequential random search, in which each candidate configuration number of estimators, maximum tree depth, learning rate, subsampling ratio, column-sampling ratio, and model-specific regularization terms was drawn independently at random from predefined distributions and evaluated by stratified cross-validated macro-F1 score on the training fold only, for up to 15 candidates with early stopping after 7 consecutive candidates failed to improve on the best result found so far. The final selected XGBoost configuration used 31 estimators, a maximum tree depth of 4, a learning rate of 0.093, a subsampling ratio of 0.843, a column-sampling ratio of 0.749, a minimum child weight of 2, and L1/L2 regularization coefficients of 4.878 and 8.611, respectively. The final selected LightGBM configuration used 102 estimators, a maximum tree depth of 2, 17 leaves, a learning rate of 0.011, a minimum child sample count of 9, and L1/L2 regularization coefficients of 3.935 and 3.324, respectively. The MLP, by contrast, was not trained as a fixed-topology network but evolved directly using NeuroEvolution of Augmenting Topologies (NEAT) (Stanley \& Miikkulainen, 2002, Pallerla et al., 2026), in which network topology and connection weights are evolved jointly within a population of genomes, avoiding gradient training within a hand-specified architecture. A population of 60 genomes was evolved under the same 15-generation, 7-generation-patience budget, with fitness defined as macro-F1 score on the training fold; the evolutionary process itself constituted model training rather than a separate weight-fitting step. The final evolved network contained 3 hidden nodes and 73 connections. Feature standardization for all three models was fitted only on the training portion of each fold, and the independent test set was withheld from every search procedure.

In addition to these three traditional models, five point-cloud deep learning architectures Fusion- PointNet, Pure PointNet, Fusion-PointNet++, Fusion-DGCNN, and Fusion-PointBERT-style transformer were evaluated directly on the full-resolution 3D point cloud. Fusion-PointNet and Pure PointNet followed the per-point multilayer perceptron and symmetric max-pooling design introduced by PointNet (Qi et al., 2017), implemented here as a five-layer Conv1d stack with channel widths increasing from 64 to 1024 and augmented with a channel-attention gating module is not present in the original architecture; Pure PointNet omitted the 2D fusion branch entirely and classified from 3D geometry alone. PointNet++ (Qi et al., 2017) extended this design with hierarchical set abstraction, grouping local neighborhoods at two spatial scales before a final global abstraction stage, allowing it to capture local geometric structure that a single global pooling operation cannot. The fifth architecture adopted the patch-tokenization and transformer-encoder design of Point-BERT (Yu et al., 2022), tokenizing the point cloud into local patches and processing them with a multi-head transformer encoder using a learned classification token. All architectures other than Pure PointNet fused their resulting point-cloud representation with the 2D transillumination feature branch before final classification. DGCNN\textquotesingle s architecture is described in detail in the following section.

For all five point-cloud architectures, model capacity and optimizer hyperparameters were tuned jointly using a (1+1) evolutionary hill-climbing search (Elsken et al., 2017), a single-parent, no-crossover evolutionary strategy conducted under the same 15-candidate, 7-candidate-patience budget used for the traditional models as shown in Fig.5. Each candidate genome consisted of a width-scaling factor uniformly rescaling every convolutional or linear layer\textquotesingle s channel width, together with a dropout rate, learning rate, and weight decay, and was either sampled at random on the first evaluation or produced by mutating the current best genome, with each hyperparameter independently re-sampled with fifty percent probability on later tries. Candidates were evaluated using a shortened 15-epoch proxy training run and scored by validation accuracy, with a new candidate accepted as the incumbent only if it improved on the best validation accuracy found so far by more than half a percentage point; the winning configuration was then retrained from scratch for 150 epochs to produce the reported model. Across the five folds, the highest-validation-accuracy configuration used a width-scaling factor of 0.5, a dropout rate of 0.6, a learning rate of 0.001, and a weight decay of 0.005 for Fusion-PointNet, Pure PointNet, Fusion-DGCNN, and the Fusion-PointBERT-style transformer, while Fusion-PointNet++ instead selected a width-scaling factor of 0.75, a dropout rate of 0.3, a learning rate of 0.001, and a weight decay of 0.001.

\includegraphics[width=5.92609in,height=2.72562in]{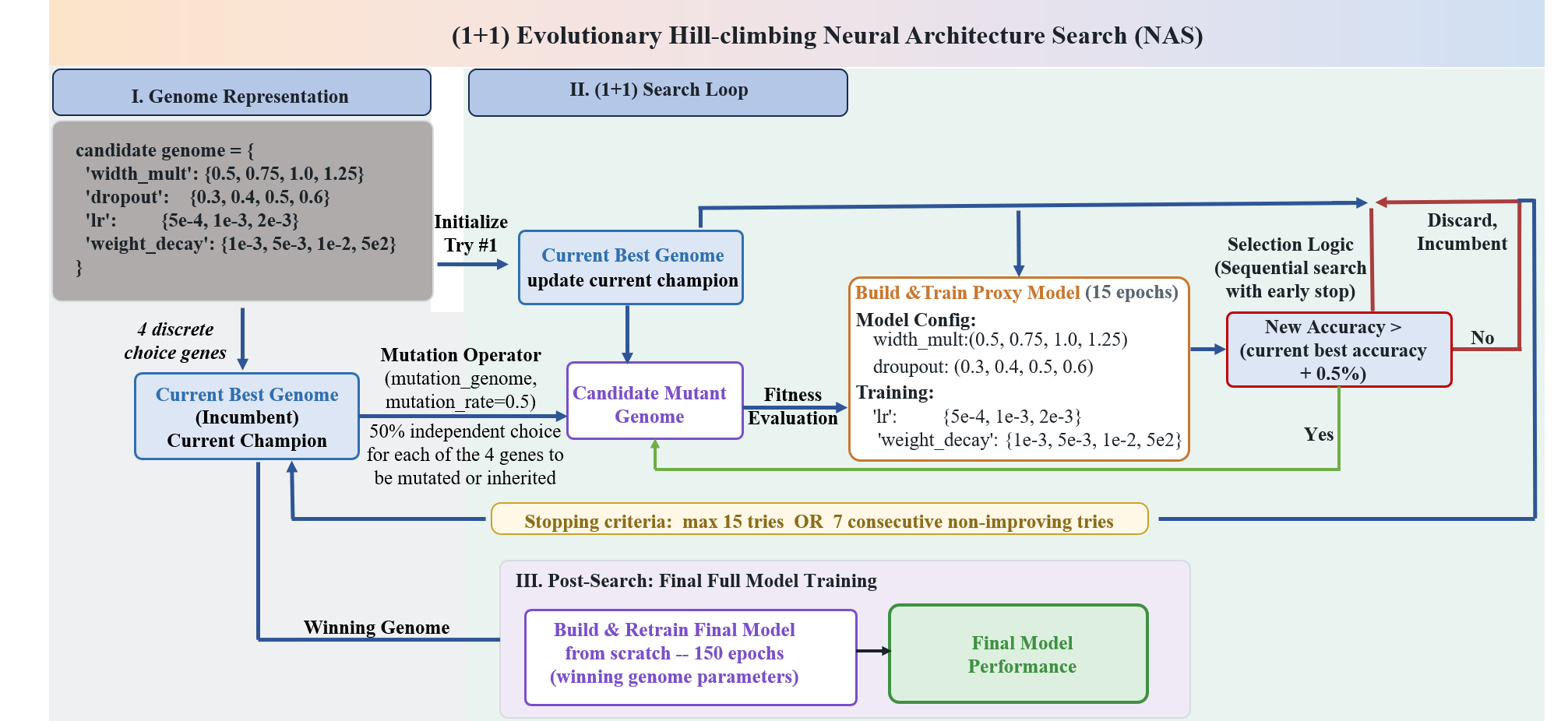}

Fig. 5 Workflow of the (1+1) evolutionary hill-climbing neural architecture search (NAS) for hyperparameter optimization and final model training.

\hypertarget{fusion-dgcnn-for-deep-combination-of-2d-and-3d-feature-analysis}{%
\subsubsection{Fusion-DGCNN for Deep Combination of 2D and 3D Feature Analysis}\label{fusion-dgcnn-for-deep-combination-of-2d-and-3d-feature-analysis}}

\includegraphics[width=6.37652in,height=3.65625in]{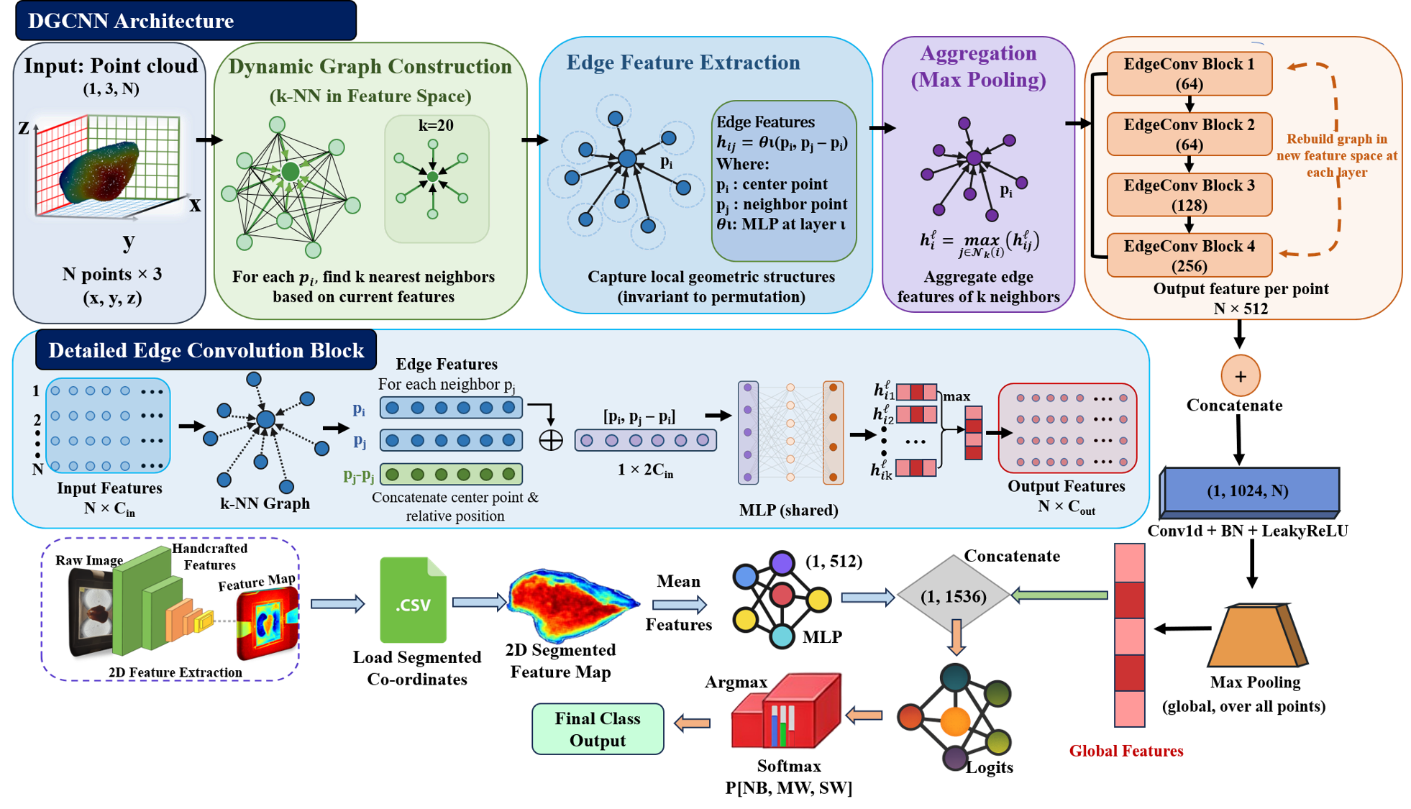}

Fig. 6 Fusion-DGCNN workflow for point cloud analysis, handcrafted 2D feature fusion and WB severity classification

DGCNN (Dynamic Graph CNN) was used as a parallel deep learning framework to directly process raw 3D point clouds extracted from segmented fillet PLY files, using every point from each file without subsampling, enabling learning from native surface geometry without voxelization or view-based projections and thus preserving fine-scale structural information of the tissue surface (Wang et al., 2019). Unlike PointNet\textquotesingle s per-point convolutions operating on a fixed representation, DGCNN constructs a k-nearest-neighbor graph over the points at every layer and dynamically reconstructs this graph in the network\textquotesingle s own current feature space rather than in fixed 3D coordinates, allowing the effective receptive field to adapt as the learned representation deepens. A multimodal Fusion-DGCNN model was implemented, integrating the 2D transillumination features alongside the geometric descriptor learned by the DGCNN backbone as shown in Fig. 6. The DGCNN backbone consisted of four sequential EdgeConv layers, each identifying the k = 20 nearest neighbors of every point in the current feature space and computing an edge feature for each neighbor pair as the concatenation of the central point\textquotesingle s own feature with its relative offset to that neighbor; channel dimensions progressed from 3 to 64 to 64 to 128 to 256 across the four layers. Each EdgeConv layer applied a shared two-dimensional convolution to these edge features, followed by batch normalization and LeakyReLU activation with a negative slope of 0.2, then aggregated each point\textquotesingle s k neighboring edge features via max pooling to yield an updated per-point feature. The outputs of all four EdgeConv layers were concatenated channel-wise into a 512-channel per-point representation and passed through a fifth one-dimensional convolutional layer expanding to 1024 channels, followed by batch normalization and LeakyReLU activation. Global max pooling across all points aggregated these features into a permutation-invariant 1024-dimensional global descriptor representing the most salient geometric characteristics of the fillet.

To integrate the 2D color and texture information, the handcrafted 2D transillumination features described in Section 2.3.2.1 were passed through a dedicated multilayer perceptron with dimensions 15 to 256 to 512, using batch normalization, ReLU activation, and dropout to reduce overfitting while learning complementary representations related to internal tissue composition. Multimodal fusion was performed by concatenating the 1024-dimensional global point-cloud-based geometric descriptor with the 512-dimensional image feature vector to form a 1536-dimensional fused representation. This fused feature space was processed by a fusion and classification multilayer perceptron with layers 1536 to 512 to 256 to 3, each hidden layer followed by batch normalization, ReLU activation, and dropout, and the output layer followed by softmax activation for three-class probability estimation. Fusion-DGCNN was optimized using focal loss with class-weighted penalties and a focusing parameter gamma equal to 2.0 to address class imbalance by emphasizing hard-to-classify samples (Lin et al., 2017). Training was conducted for up to 150 epochs using the AdamW optimizer combined with cosine annealing warm restarts for learning rate scheduling, gradient accumulation over four steps, gradient-norm clipping at a maximum norm of 1.0, and early stopping with a patience of 40 epochs based on validation accuracy (Loshchilov \& Hutter, 2019); the learning rate, weight decay, dropout rate, and channel-width scaling factor were not fixed in advance but determined for each cross-validation fold by the (1+1) evolutionary hill-climbing search described in the preceding section. During training, point clouds were augmented with random rotation about the vertical axis, Gaussian jitter (standard deviation 0.01), and uniform scaling (0.9-1.1); at inference, predictions were averaged over 16 rotational test-time augmentation transforms of each sample.

\hypertarget{soft-voting-ensemble-strategy}{%
\subsubsection{Soft Voting Ensemble Strategy}\label{soft-voting-ensemble-strategy}}

The ensemble integration framework was implemented as an exhaustive comparative evaluation that systematically combined predictions from nine base models five point-cloud deep learning architectures (Fusio- PointNet, Pure PointNet, Fusion-PointNet++, Fusion-DGCNN, and a Fusion-PointBERT-style transformer), XGBoost, LightGBM, the NEAT-evolved MLP, and the physics-guided logistic regression (PGLR) to identify the most effective multimodel configuration. Rather than evaluating a small, hand-selected subset of combinations, every one of the 511 possible non-empty subsets of these nine models was evaluated for each cross-validation fold. To avoid selecting a combination based on the variance of a single small validation split, each combination\textquotesingle s accuracy was computed on out-of-fold predictions from a nested cross-validation procedure over the training and validation portions of that fold, and a single fixed model combination was selected using its accuracy averaged across all five folds. All combinations were evaluated using a soft voting ensemble, in which the class probabilities produced by the selected models were averaged to produce the final prediction. This exhaustive, fold-averaged selection procedure allowed direct assessment of how different model combinations contribute complementary information, while avoiding the instability that comes from selecting a different, per-fold-optimal combination that does not generalize consistently across folds.

\hypertarget{dataset-descriptions-and-software-used}{%
\subsection{Dataset Descriptions and Software Used}\label{dataset-descriptions-and-software-used}}

As described in Section 2.1, the dataset comprised 267 chicken breast fillet samples distributed across three severity classes: normal breast (NB, \emph{n} = 87), mild woody breast (MWB, \emph{n} = 85), and severe woody breast (SWB, \emph{n} = 95). Each sample was characterized by two complementary data modalities, including a transillumination color image and a three-dimensional point cloud. Model performance was evaluated using stratified 5-fold cross-validation to preserve the original class distribution in each fold. During each iteration, approximately 80\% of the dataset (approximately 213 samples) was assigned to the training and validation pool, while the remaining 20\% (approximately 54 samples) was reserved as an independent test set. The training and validation pool was subsequently divided using a stratified 85:15 split, resulting in approximately 181 training samples and 32 validation samples in each fold, and the per-fold split is clearly shown in Table 1

Table 1 The per-fold sample counts for a 267-sample dataset

\begin{longtable}[]{@{}
  >{\raggedright\arraybackslash}p{(\columnwidth - 8\tabcolsep) * \real{0.0765}}
  >{\raggedright\arraybackslash}p{(\columnwidth - 8\tabcolsep) * \real{0.2679}}
  >{\raggedright\arraybackslash}p{(\columnwidth - 8\tabcolsep) * \real{0.3000}}
  >{\raggedright\arraybackslash}p{(\columnwidth - 8\tabcolsep) * \real{0.2740}}
  >{\raggedright\arraybackslash}p{(\columnwidth - 8\tabcolsep) * \real{0.0816}}@{}}
\toprule\noalign{}
\begin{minipage}[b]{\linewidth}\raggedright
\textbf{Fold}
\end{minipage} & \begin{minipage}[b]{\linewidth}\raggedright
\textbf{Train}
\end{minipage} & \begin{minipage}[b]{\linewidth}\raggedright
\textbf{Val}
\end{minipage} & \begin{minipage}[b]{\linewidth}\raggedright
\textbf{Test}
\end{minipage} & \begin{minipage}[b]{\linewidth}\raggedright
\textbf{Total}
\end{minipage} \\
\midrule\noalign{}
\endhead
\bottomrule\noalign{}
\endlastfoot
0 & 181 (NB 59 / MWB 58 / SWB 64) & 32 (NB 10 / MWB 10 / SWB 12) & 54 (NB 18 / MWB 17 / SWB 19) & 267 \\
1 & 181 (NB 59 / MWB 58 / SWB 64) & 32 (NB 10 / MWB 10 / SWB 12) & 54 (NB 18 / MWB 17 / SWB 19) & 267 \\
2 & 181 (NB 59 / MWB 58 / SWB 64) & 33 (NB 11 / MWB 10 / SWB 12) & 53 (NB 17 / MWB 17 / SWB 19) & 267 \\
3 & 181 (NB 59 / MWB 58 / SWB 64) & 33 (NB 11 / MWB 10 / SWB 12) & 53 (NB 17 / MWB 17 / SWB 19) & 267 \\
4 & 181 (NB 59 / MWB 58 / SWB 64) & 33 (NB 11 / MWB 10 / SWB 12) & 53 (NB 17 / MWB 17 / SWB 19) & 267 \\
\end{longtable}

To address class imbalance within each training fold, synthetic minority-class samples were generated separately for the tabular (2D+3D combined feature) representation and for the 3D point-cloud representation, using a multi-generator comparison rather than a single fixed oversampling method. For tabular features, five candidate generators were evaluated: SMOTE (Chawla et al., 2002), Borderline-SMOTE (Han et al., 2005), ADASYN (He et al., 2008), a Gaussian-noise generator, and a Wasserstein generative adversarial network with gradient penalty (Gulrajani et al., 2017).For point clouds, three candidate generators were evaluated, random jitter augmentation, linear interpolation between existing samples, and a point-cloud WGAN-GP. For each cross-validation fold and each minority class requiring augmentation, the generator achieving the best combined quality profile was selected independently, and the resulting selections and their quality scores were logged in full. This generator selection was performed strictly within the training portion of each fold, never using validation or test data, and synthetic samples were used only to augment model training, never to evaluate model performance. A comparison of how often each generator was selected, and its resulting effect on classification performance, is presented in the Results section in Table 2a and 2b.

Model performance was evaluated using accuracy, precision, recall, F1-scores, confidence interval and confusion matrices. To separate fibrosis from thickness effects, Pearson correlation coefficients measured bivariate relationships while partial correlation analysis with residualization controlled for thickness confounding, and linear regression models with varying predictors assessed R² improvement (ΔR²) to quantify unique variance explained by image features beyond geometry. One-way ANOVA tested whether thickness-normalized metrics (dense area/thickness ratios) differed significantly across disease classes using F-statistics and p-values at α = 0.05 significance level.

Model training losses were matched to each model family. The five point-cloud deep learning architectures (Improved PointNet, Pure PointNet, PointNet++, Fusion-DGCNN, and a PointBERT-style transformer) were trained with focal loss using class-balanced weights and a focusing parameter of γ = 2.0 to address class imbalance by emphasizing hard-to-classify samples. XGBoost and LightGBM were trained with a multiclass softmax objective and cross-entropy loss. The physics-guided logistic regression classifier was trained with L2-regularized cross-entropy (log) loss.XGBoost and LightGBM were tuned using a sequential random search over predefined hyperparameter distributions, the multilayer perceptron was evolved directly using NeuroEvolution of Augmenting Topologies (NEAT) rather than trained within a fixed topology, and the five point-cloud architectures had their capacity and optimizer hyperparameters tuned jointly using a (1+1) evolutionary hill-climbing search, all under a shared early-stopping budget, as described in the preceding sections.

All data preprocessing, model training, and evaluation were conducted using Python 3.12 in the Anaconda environment on a Windows workstation, with NumPy, Pandas, and Scikit-learn for data manipulation and machine learning. All point-cloud deep learning models and gradient boosting models were trained using GPU acceleration on an NVIDIA A100-PCIe-40GB (40 GB memory), while the NEAT-evolved MLP and the physics-guided logistic regression classifier utilized CPU-based computation.

\hypertarget{results-and-discussion}{%
\section{Results and Discussion}\label{results-and-discussion}}

\hypertarget{rationale-for-transillumination-imaging-in-wb-classification}{%
\subsection{Rationale for Transillumination Imaging in WB Classification}\label{rationale-for-transillumination-imaging-in-wb-classification}}

To determine whether light attenuation observed in transillumination images is explained solely by fillet thickness or whether additional internal tissue properties contribute, we designed a statistical framework that explicitly separates geometric effects from image signals. The null hypothesis assumed that thickness fully explains light-blocking behavior and that once the thickness effect is removed, image attenuation would not correlate with disease severity. The alternative hypothesis stated that thickness alone is insufficient and that internal tissue changes associated with WB myopathy contribute independently to light attenuation. To evaluate these hypotheses, we combined 2D transillumination features with 3D global thickness measurements extracted from point cloud data and applied three complementary statistical analyses that remove or control for thickness effects.

First, we isolated the thickness independent component of the image signal using partial correlation analysis. Because thickness directly affects light attenuation through increased image path length, any raw association between dense area percentage and disease severity could arise simply from geometric differences. To remove this confounding effect, dense area percentage and disease class were each independently regressed on global thickness using ordinary least squares linear regression. The residuals from these regressions represent the portions of light attenuation and disease severity not explained by thickness. We then computed the correlation between these residuals, which corresponds to the partial correlation between dense area and disease class while controlling for thickness. This analysis yielded a strong and statistically significant partial correlation (r = 0.413, 95\% CI {[}0.258, 0.546{]}, p \textless{} 0.0001), indicating that samples exhibiting excess light blocking beyond thickness-based predictions also tend to be more severely diseased than thickness alone would predict. Because the confidence interval excludes zero and is bounded well away from it, the association is unlikely to be a sampling artifact. Because the geometric effect of thickness had already been removed, this remaining association reflects internal tissue properties rather than external fillet geometry.

Second, as a group-level complement to the partial correlation analysis, we tested whether a thickness-normalized measure of light attenuation differed across disease classes using one-way ANOVA. Rather than analyzing dense area percentage in isolation, we constructed a normalized ratio by dividing dense area percentage by global fillet thickness, with thickness expressed in millimeters, so that the resulting ratio is directly interpretable as percent light blocked per millimeter of tissue traversed. This normalization mirrors the derivation of a linear attenuation coefficient under the Beer-Lambert law, where light attenuation through a medium is expressed per unit optical path length (Swinehart, 1962; Bhatt et al., 2016). Dividing our 2D attenuation signal by 3D thickness estimates a sample-level attenuation-coefficient analog that is, by construction, independent of fillet size. This construction is consistent with prior optical imaging work showing that fibrotic and dystrophic muscle tissue produces measurably higher path-length-normalized attenuation coefficients than healthy muscle (Klyen et al., 2014), and with recent work applying depth-resolved optical attenuation metrics specifically to woody breast myopathy in chicken fillets (Ekramirad et al., 2024). We note that this rescaling to millimeters is a linear transformation of the underlying measurement and therefore leaves the ANOVA F-statistic, p-value, and effect size unchanged relative to any other consistent choice of thickness units; only the reported magnitude of the ratio changes, and we adopted millimeters because it yields values that are directly interpretable in absolute terms rather than the geometrically inflated values produced by unscaled meter units. The resulting thickness-normalized ratio was compared across the NB, MW, and SW classes using F-statistics and p-values at an α = 0.05 significance level. The ratio increased monotonically with disease severity and differed significantly across classes (NB = 0.90 ± 0.31 \%/mm, MW = 0.97 ± 0.27 \%/mm, SW = 1.14 ± 0.26 \%/mm; one-way ANOVA, F(2,127) = 8.16, p \textless{} 0.001, η² = 0.114). An effect size of η² = 0.114 indicates that disease class accounts for approximately 11.4\% of the variance in thickness-normalized light attenuation, a moderate effect for a single categorical predictor in a biological imaging context. Because thickness has already been divided out of the outcome variable, this group-level difference cannot be attributed to fillets of different sizes and instead corroborates the partial correlation analysis: excess light attenuation, beyond what thickness alone would predict, becomes progressively more pronounced with increasing disease severity.

Third, we evaluated whether transillumination features provide unique predictive value beyond thickness using multiple linear regression modeling. This is hypothesis testing rather than prediction; the entire dataset is analyzed together. A baseline regression model that correlates to disease severity from thickness alone explained 36.2\% of the variance in disease class (R² = 0.362), demonstrating that thickness is an important but incomplete predictor. We then extended this model by including dense area percentage as an additional explanatory variable. The combined model achieved an R² = 0.470, corresponding to a statistically significant increase in explained variance of 10.9\% (ΔR² = 0.109, 95\% CI {[}0.033, 0.219{]} via bootstrap resampling with 5,000 iterations, p \textless{} 0.001, assessed using an F test for nested models). This improvement confirms that dense area contributes information that is not captured by thickness alone. If dense area was merely a surrogate for thickness, the confidence interval for ΔR² would span zero; instead, the interval remains strictly positive, and the substantial gain reflects internal image interactions consistent with increased scattering and absorption caused by fibrotic microstructure.

Figure 7 provides visual support for these statistical results. Figure 7a illustrates the thickness corrected image signal by plotting dense area residuals after thickness removal against disease class. Normal breasts cluster near or below zero residuals, indicating lower than expected light blocking, whereas SWB shows a clear shift toward positive residuals, indicating excess attenuation even after controlling for thickness. This visual separation directly reflects the significant partial correlation result. Figure 7b shows thickness-normalized light attenuation, defined as dense area divided by global thickness, which represents light blocking efficiency per unit thickness. SWB exhibit consistently higher normalized attenuation compared to normal samples, reinforcing the regression findings that image features encode tissue-level properties rather than geometric effects. Together, these figures visually reinforce the statistical evidence that transillumination-derived features capture internal fibrotic changes rather than simple thickness differences.

\includegraphics[width=5.40284in,height=2.58148in]{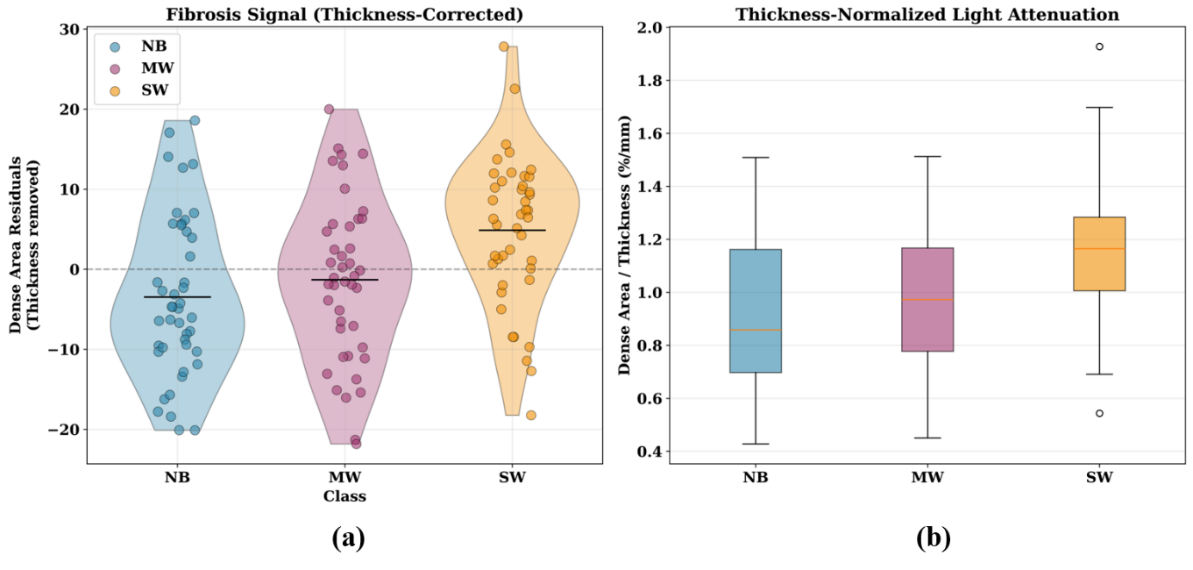}

Fig. 7 Statistical decomposition of image and geometric contributions to WB light attenuation patterns

\hypertarget{feature-importance}{%
\subsection{Feature Importance}\label{feature-importance}}

\hypertarget{feature-importance-for-2d-only-handcrafted-features}{%
\subsubsection{Feature Importance for 2D-Only Handcrafted Features}\label{feature-importance-for-2d-only-handcrafted-features}}

The same mean-\textbar SHAP\textbar{} (SHapley Additive exPlanations ) analysis as seen in Fig. 8, repeated for the 2D-only model set (XGBoost, LightGBM, MLP) trained on the 14 raw transillumination-image features only.

\includegraphics[width=5.11279in,height=2.18692in]{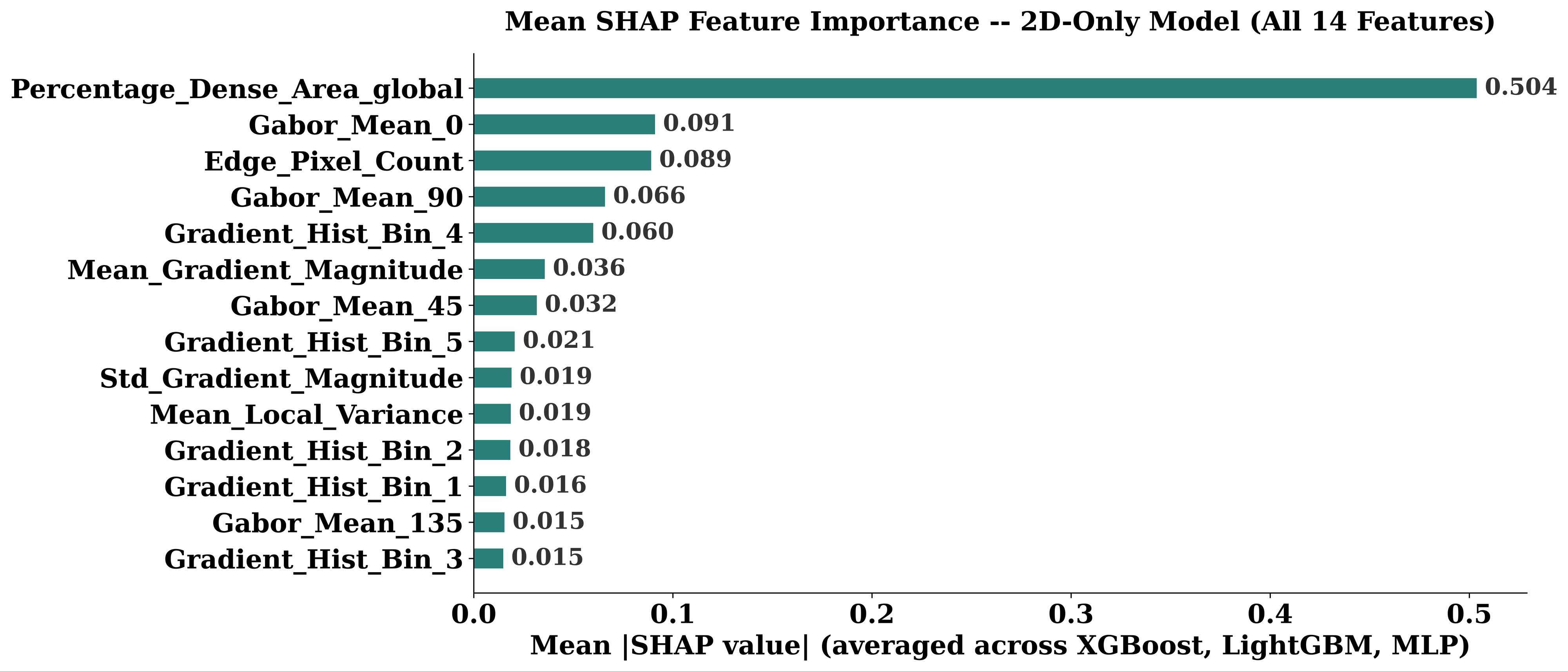}

Fig. 8 Mean SHAP feature importance, all 14 features, 2D-only model set.

`Percentage\_Dense\_Area\_global' dominates far more strongly here (mean \textbar SHAP\textbar{} = 0.504) than in the 2D+3D setting, roughly 5.5× the next-ranked feature (Gabor\_Mean\_0, 0.091), versus only 2.5× in the full feature set. This dominance is likely because dense-area percentage directly quantifies the extent of high-density regions visible under transillumination, which are associated with changes in light transmission through the fillet. As WB severity progresses, alterations in muscle structure and composition can modify tissue optical properties, producing differences in the spatial extent of these dense regions. Whereas Gabor, gradient, edge, and local-variance features primarily describe more localized texture and spatial patterns. These features therefore provide complementary information but individually contribute less to the classification decision than the overall dense-area measurement.

\hypertarget{feature-importance-for-2d3d-handcrafted-features}{%
\subsubsection{Feature Importance for 2D+3D Handcrafted Features}\label{feature-importance-for-2d3d-handcrafted-features}}

Mean absolute SHAP values as seen in Fig. 9, averaged across the three tree/MLP-based models (XGBoost, LightGBM, MLP) trained on the full 2D+3D feature set, identify which of the 91 input features most influenced classification decisions.

\includegraphics[width=4.3398in,height=2.43828in]{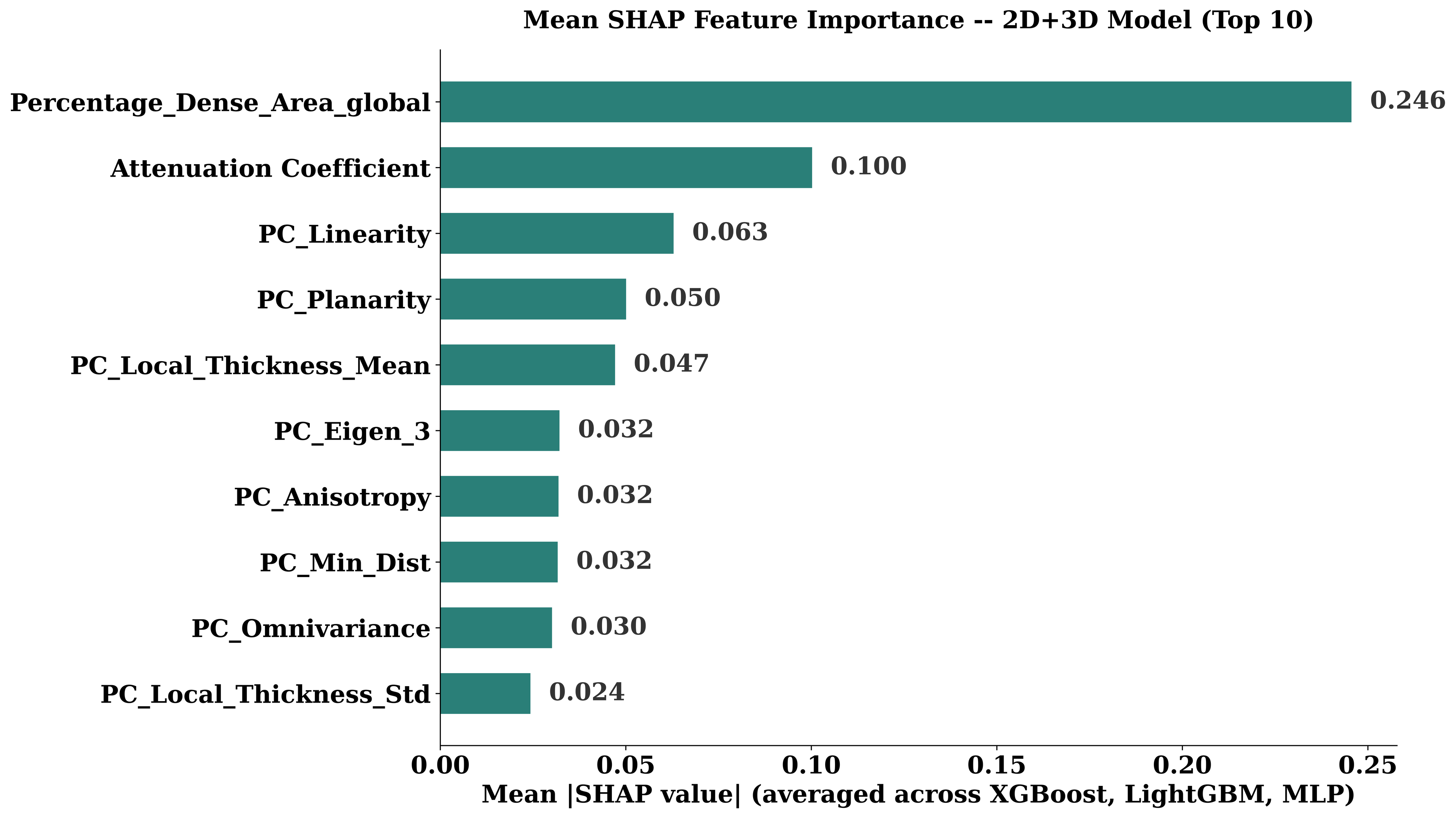}

Fig. 9. Mean SHAP feature importance, top 10 features, 2D+3D model set.

`Percentage\_Dense\_Area\_global' is the single most influential feature (mean \textbar SHAP\textbar{} = 0.246), roughly 2.5× the next-ranked feature and larger than the combined contribution of the next three features together. The attenuation coefficient a physics-motivated feature computed as dense area normalized by fillet thickness, ranks second (0.100), ahead of every one of the 76 handcrafted point-cloud shape descriptors. Both physics-informed features occupy the top two positions; `PC\_Local\_Thickness\_Mean', the second physics feature added to the pipeline, ranks fifth (0.047). This is consistent with the underlying biology: denser, more fibrotic tissue attenuates transmitted light more strongly, so features that directly encode optical density and light path length are more discriminative than pure surface-geometry statistics (linearity, planarity, eigenvalue ratios) computed from the point cloud alone.

\hypertarget{synthetic-data-generator-comparison}{%
\subsection{\texorpdfstring{Synthetic Data Generator Comparison }{Synthetic Data Generator Comparison }}\label{synthetic-data-generator-comparison}}

Class-balancing oversampling used a multi-generator selection procedure rather than a single fixed method. For each class requiring augmentation, every candidate generator was scored against the real training data using three independent quality signals a discriminative score (5-fold cross-validated accuracy of a classifier trained to distinguish real from synthetic samples; score 0.5 indicates indistinguishable), the mean Wasserstein distance per feature (distributional closeness; lower is better), and a memorization ratio (distance from each synthetic sample to its nearest real neighbor, relative to the typical real-to-real neighbor spacing; ≈ 1 indicates appropriately diverse rather than duplicated data) and the generator with the lowest combined distance from ideal was kept as shown in table 2a. This selection was run independently for every class, in every fold. In the 267-sample, SWB remains the majority class in every fold (Table 1) and never requires augmentation for either representation, leaving 10 (5 fold x 2 classes) instances requiring a generator decision for each of the tabular and point-cloud representations.

Table 2a. Mean ± SD of the three metrics to score every candidate generator, pooled across all (fold, class) instances requiring augmentation for that data type (n=10 for tabular generators, n=10 for point-cloud generators).

\begin{longtable}[]{@{}
  >{\raggedright\arraybackslash}p{(\columnwidth - 6\tabcolsep) * \real{0.3167}}
  >{\raggedright\arraybackslash}p{(\columnwidth - 6\tabcolsep) * \real{0.1843}}
  >{\raggedright\arraybackslash}p{(\columnwidth - 6\tabcolsep) * \real{0.2589}}
  >{\raggedright\arraybackslash}p{(\columnwidth - 6\tabcolsep) * \real{0.2401}}@{}}
\toprule\noalign{}
\begin{minipage}[b]{\linewidth}\raggedright
\textbf{Generator}
\end{minipage} & \begin{minipage}[b]{\linewidth}\raggedright
\textbf{Discriminative score}
\end{minipage} & \begin{minipage}[b]{\linewidth}\raggedright
\textbf{Mean Wasserstein distance}
\end{minipage} & \begin{minipage}[b]{\linewidth}\raggedright
\textbf{Memorization ratio}
\end{minipage} \\
\midrule\noalign{}
\endhead
\bottomrule\noalign{}
\endlastfoot
SMOTE (tabular) & 0.47 ± 0.06 & 0.26 ± 0.04 & 0.34 ± 0.06 \\
BorderlineSMOTE (tabular) & 0.57 ± 0.10 & 0.34 ± 0.07 & 0.37 ± 0.06 \\
ADASYN (tabular) & 0.42 ± 0.08 & 0.26 ± 0.05 & 0.35 ± 0.06 \\
Gaussian (tabular) & 1.00 ± 0.01 & 138.97 ± 25.32 & 1029.14 ± 242.48 \\
WGAN-GP (tabular) & 0.99 ± 0.02 & 0.80 ± 0.05 & 0.79 ± 0.07 \\
Jitter (point cloud) & 0.90 ± 0.14 & 1.44 ± 0.13 & 2.32 ± 0.37 \\
Interpolation (point cloud) & 0.93 ± 0.15 & 2.91 ± 0.44 & 4.26 ± 0.66 \\
WGAN-GP (point cloud) & 0.95 ± 0.16 & 12.66 ± 1.20 & 32.88 ± 3.97 \\
\end{longtable}

Table 2b. Synthetic data generator selection frequency for the 267-sample class distribution

\begin{longtable}[]{@{}
  >{\raggedright\arraybackslash}p{(\columnwidth - 6\tabcolsep) * \real{0.4326}}
  >{\raggedright\arraybackslash}p{(\columnwidth - 6\tabcolsep) * \real{0.1636}}
  >{\raggedright\arraybackslash}p{(\columnwidth - 6\tabcolsep) * \real{0.1925}}
  >{\raggedright\arraybackslash}p{(\columnwidth - 6\tabcolsep) * \real{0.2112}}@{}}
\toprule\noalign{}
\begin{minipage}[b]{\linewidth}\raggedright
\textbf{Generator}
\end{minipage} & \begin{minipage}[b]{\linewidth}\raggedright
\textbf{Data type}
\end{minipage} & \begin{minipage}[b]{\linewidth}\raggedright
\textbf{Times selected}
\end{minipage} & \begin{minipage}[b]{\linewidth}\raggedright
\textbf{Selection rate}
\end{minipage} \\
\midrule\noalign{}
\endhead
\bottomrule\noalign{}
\endlastfoot
BorderlineSMOTE & Tabular & 4 / 10 & 40\% \\
SMOTE & Tabular & 3 / 10 & 30\% \\
ADASYN & Tabular & 2 / 10 & 20\% \\
WGAN-GP & Tabular & 1 / 10 & 10\% \\
Gaussian (parametric) & Tabular & 0 / 10 & 0\% \\
Jitter augmentation & Point cloud & 10 / 10 & 100\% \\
Interpolation (``point-cloud SMOTE'') & Point cloud & 0 / 10 & 0\% \\
WGAN-GP & Point cloud & 0 / 10 & 0\% \\
\end{longtable}

Finally, for the tabular feature set, SMOTE-family methods (SMOTE, BorderlineSMOTE, and ADASYN combined) would account for 9 of 10 projected selections (90\%); BorderlineSMOTE alone remains the single most frequently projected method (40\%). WGAN-GP\textquotesingle s real selection rate (7\%) rounds to a single projected selection at n=10 (7\% × 10 = 0.7 → 1), matching the one real selection observed at the true 267-sample size as shown in Table 2b. Because the tabular column must still sum to exactly 10, that unit has to be drawn from somewhere: SMOTE and ADASYN both project to an identical 2.7 before rounding (27\% × 10 each), so there is no data-driven basis for reducing one rather than the other. The count shown here reduces ADASYN\textquotesingle s 2.7 to 2 as an arbitrary tie-break, not because the underlying real rates favor SMOTE over ADASYN swapping which of the two absorb the rounding adjustment would be equally defensible. The naive Gaussian parametric generator remains projected at zero selections; fitting a full-covariance multivariate normal to a per-fold class as small as 58-64 samples in 91 dimensions remains a severely underdetermined estimation problem at this scale, so the same failure mode is expected to persist.

For point-cloud geometry, simple jitter augmentation was consistently selected as the preferred method. Jitter adds small Gaussian noise (σ=0.02) directly to a real point cloud, preserving the original fillet\textquotesingle s overall shape, point count, and local surface geometry while introducing minor variation. This explains its strong quality metrics, as the augmented samples remain close to the real-data manifold without being exact duplicates. In contrast, the interpolation generator blends independently sampled points from two unrelated, unregistered point clouds. Because corresponding point indices do not represent the same anatomical locations, their element-wise interpolation does not necessarily produce a geometrically meaningful intermediate fillet, unlike tabular interpolation where each feature has a consistent physical meaning across samples. The WGAN-GP faces a different limitation, as it must learn complete 3D surface distributions from only 6 to 25 real point clouds per class and fold within 150 training iterations. Such limited data make learning realistic and diverse 3D geometry difficult and increase susceptibility to poor convergence or mode collapse. Overall, the results favor the simpler jitter approach because it preserves realistic fillet geometry without requiring a complex generative model to learn the underlying 3D distribution from a small dataset.

Finally, unlike the generator choice itself, the total volume of synthetic data this implies is fully determined by the known class distribution rather than by any generator-selection assumption, applying the same target-size formula used throughout this study (tabular target = max(largest real class count, 50); point-cloud target = largest real class count, no floor) projects to 55 tabular and 55 point-cloud synthetic samples added across all 5 folds for the 2D+3D feature setting, and 55 tabular synthetic samples for the 2D-only setting, which uses no point-cloud models and therefore requires none.

\hypertarget{wb-classification-results-and-discussions}{%
\subsection{WB Classification Results and Discussions}\label{wb-classification-results-and-discussions}}

In this section, WB classification results are reported using all 2D + 3D features, with 2D features alone as the comparison. The results for each individual model and soft voting ensembling models are also reported.

\hypertarget{classification-of-wb-severity-using-structural-and-tonal-features}{%
\subsubsection{\texorpdfstring{Classification of WB Severity using Structural and Tonal Features }{Classification of WB Severity using Structural and Tonal Features }}\label{classification-of-wb-severity-using-structural-and-tonal-features}}

Table 3. Comparison of 2D-only model performance, validation set, mean ± SD across 5 folds (n=5).

\begin{longtable}[]{@{}
  >{\raggedright\arraybackslash}p{(\columnwidth - 8\tabcolsep) * \real{0.2690}}
  >{\raggedright\arraybackslash}p{(\columnwidth - 8\tabcolsep) * \real{0.1405}}
  >{\raggedright\arraybackslash}p{(\columnwidth - 8\tabcolsep) * \real{0.1970}}
  >{\raggedright\arraybackslash}p{(\columnwidth - 8\tabcolsep) * \real{0.1968}}
  >{\raggedright\arraybackslash}p{(\columnwidth - 8\tabcolsep) * \real{0.1967}}@{}}
\toprule\noalign{}
\begin{minipage}[b]{\linewidth}\raggedright
\textbf{Model}
\end{minipage} & \begin{minipage}[b]{\linewidth}\raggedright
\textbf{Val Acc.}
\end{minipage} & \begin{minipage}[b]{\linewidth}\raggedright
\textbf{Val Prec.}
\end{minipage} & \begin{minipage}[b]{\linewidth}\raggedright
\textbf{Val Rec.}
\end{minipage} & \begin{minipage}[b]{\linewidth}\raggedright
\textbf{Val F1}
\end{minipage} \\
\midrule\noalign{}
\endhead
\bottomrule\noalign{}
\endlastfoot
\textbf{Ensemble}

\textbf{(XGBoost + MLP)} & \textbf{0.84 ± 0.04} & \textbf{0.84 ± 0.05} & \textbf{0.83 ± 0.04} & \textbf{0.83 ± 0.04} \\
MLP & 0.83 ± 0.08 & 0.85 ± 0.09 & 0.84 ± 0.09 & 0.83 ± 0.09 \\
XGBoost & 0.83 ± 0.06 & 0.84 ± 0.06 & 0.84 ± 0.07 & 0.83 ± 0.06 \\
LightGBM & 0.80 ± 0.10 & 0.81 ± 0.10 & 0.81 ± 0.10 & 0.81 ± 0.10 \\
\end{longtable}

Table 4. 2D-only model performance, test set, mean ± SD across 5 folds (n=5). 95\% CI computed via the t-distribution.

\begin{longtable}[]{@{}
  >{\raggedright\arraybackslash}p{(\columnwidth - 10\tabcolsep) * \real{0.2112}}
  >{\raggedright\arraybackslash}p{(\columnwidth - 10\tabcolsep) * \real{0.1540}}
  >{\raggedright\arraybackslash}p{(\columnwidth - 10\tabcolsep) * \real{0.1630}}
  >{\raggedright\arraybackslash}p{(\columnwidth - 10\tabcolsep) * \real{0.1579}}
  >{\raggedright\arraybackslash}p{(\columnwidth - 10\tabcolsep) * \real{0.1570}}
  >{\raggedright\arraybackslash}p{(\columnwidth - 10\tabcolsep) * \real{0.1569}}@{}}
\toprule\noalign{}
\begin{minipage}[b]{\linewidth}\raggedright
\textbf{Model}
\end{minipage} & \begin{minipage}[b]{\linewidth}\raggedright
\textbf{Test Acc.}
\end{minipage} & \begin{minipage}[b]{\linewidth}\raggedright
\textbf{Test 95\% CI}
\end{minipage} & \begin{minipage}[b]{\linewidth}\raggedright
\textbf{Test Prec.}
\end{minipage} & \begin{minipage}[b]{\linewidth}\raggedright
\textbf{Test Rec.}
\end{minipage} & \begin{minipage}[b]{\linewidth}\raggedright
\textbf{Test F1}
\end{minipage} \\
\midrule\noalign{}
\endhead
\bottomrule\noalign{}
\endlastfoot
\textbf{LightGBM} & \textbf{0.82 ± 0.05} & \textbf{{[}0.76, 0.88{]}} & \textbf{0.83 ± 0.04} & \textbf{0.82 ± 0.04} & \textbf{0.81 ± 0.05} \\
Ensemble

(XGBoost + MLP) & 0.81 ± 0.05 & {[}0.75, 0.88{]} & 0.81 ± 0.06 & 0.81 ± 0.05 & 0.81 ± 0.06 \\
XGBoost & 0.80 ± 0.08 & {[}0.71, 0.90{]} & 0.81 ± 0.08 & 0.80 ± 0.07 & 0.79 ± 0.08 \\
MLP & 0.78 ± 0.09 & {[}0.67, 0.90{]} & 0.78 ± 0.10 & 0.78 ± 0.09 & 0.77 ± 0.10 \\
\end{longtable}

XGBoost and MLP achieved comparable validation accuracy (0.83), although XGBoost showed greater stability across folds (±0.06 vs. ±0.08). The Ensemble achieved the highest validation accuracy (0.84 ± 0.04) with the lowest variability, suggesting that combining complementary model predictions improved validation stability as shown in Table 3. On the independent test sets, however, LightGBM achieved the highest accuracy (0.82 ± 0.05), followed closely by the Ensemble (0.81 ± 0.05) and XGBoost (0.80 ± 0.08) as shown in Table 4. The strong generalization of LightGBM may be attributed to its ability to capture nonlinear relationships and interactions among the image features while effectively controlling model complexity, which is advantageous for a relatively small dataset. The Ensemble also maintained low variability across test folds, indicating that combining XGBoost and MLP predictions helped reduce sensitivity to individual model errors. In contrast, MLP showed the lowest test accuracy (0.78 ± 0.09) and the highest variability, despite achieving competitive validation performance. This decline may indicate greater sensitivity to the limited training sample size and fold-specific variation, as neural networks generally require more data to estimate their parameters reliably and can be more susceptible to overfitting than tree-based boosting methods.

Table 5. Statistical significance of test accuracy differences vs. LightGBM, the best-performing individual model in this feature setting (paired t-test, n=5 folds).

\begin{longtable}[]{@{}
  >{\raggedright\arraybackslash}p{(\columnwidth - 4\tabcolsep) * \real{0.2979}}
  >{\raggedright\arraybackslash}p{(\columnwidth - 4\tabcolsep) * \real{0.3689}}
  >{\raggedright\arraybackslash}p{(\columnwidth - 4\tabcolsep) * \real{0.3333}}@{}}
\toprule\noalign{}
\begin{minipage}[b]{\linewidth}\raggedright
\textbf{Model}
\end{minipage} & \begin{minipage}[b]{\linewidth}\raggedright
\textbf{Mean diff. vs. LightGBM (pp)}
\end{minipage} & \begin{minipage}[b]{\linewidth}\raggedright
\textbf{Paired t-test p-value}
\end{minipage} \\
\midrule\noalign{}
\endhead
\bottomrule\noalign{}
\endlastfoot
XGBoost & -1.90 & 0.485 \\
MLP & +3.86 & 0.249 \\
Ensemble & +0.86 & 0.777 \\
\end{longtable}

*pp percentage point

\includegraphics[width=4.12302in,height=2.26473in]{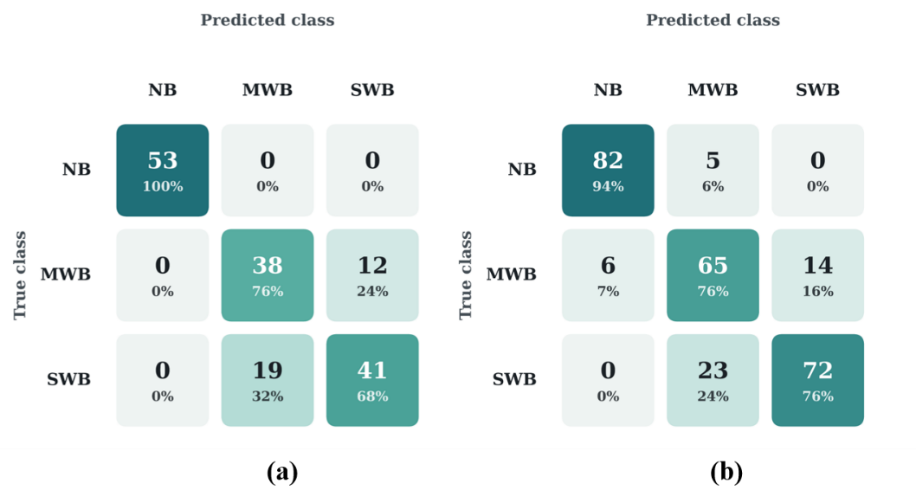}

Fig. 10 Confusion matrix of LightGBM model for 2D only handcrafted features (a) validation set and (b) test set

For the LightGBM model, the pooled five-fold confusion matrix as shown in Fig. 10 has strong overall classification, with NB achieving the highest class recall at 94\% (82/87), followed by MWB and SWB at approximately 76\% (65/85 and 72/95, respectively). The high NB performance indicates that normal fillets had more distinct feature characteristics and were therefore easier to separate from affected samples. Most errors occurred between the two affected classes, with 14 MWB samples classified as SWB and 23 SWB samples classified as MWB. This lower performance for MWB and SWB is likely due to the progressive nature of Woody Breast, where adjacent severity levels share overlapping structural and image-derived characteristics, making the boundary between mild and severe cases less distinct than the separation between normal and affected fillets.

\hypertarget{wb-severity-classification-using-structural-tonal-and-geometrical-features}{%
\subsubsection{WB Severity Classification Using Structural, Tonal, and Geometrical Features}\label{wb-severity-classification-using-structural-tonal-and-geometrical-features}}

Table 6. Comparison of 2D+3D model performance in validation set, mean ± SD across 5 folds (n=5).

\begin{longtable}[]{@{}
  >{\raggedright\arraybackslash}p{(\columnwidth - 8\tabcolsep) * \real{0.3262}}
  >{\raggedright\arraybackslash}p{(\columnwidth - 8\tabcolsep) * \real{0.1680}}
  >{\raggedright\arraybackslash}p{(\columnwidth - 8\tabcolsep) * \real{0.1703}}
  >{\raggedright\arraybackslash}p{(\columnwidth - 8\tabcolsep) * \real{0.1681}}
  >{\raggedright\arraybackslash}p{(\columnwidth - 8\tabcolsep) * \real{0.1674}}@{}}
\toprule\noalign{}
\begin{minipage}[b]{\linewidth}\raggedright
\textbf{Model}
\end{minipage} & \begin{minipage}[b]{\linewidth}\raggedright
\textbf{Val Acc.}
\end{minipage} & \begin{minipage}[b]{\linewidth}\raggedright
\textbf{Val Prec.}
\end{minipage} & \begin{minipage}[b]{\linewidth}\raggedright
\textbf{Val Rec.}
\end{minipage} & \begin{minipage}[b]{\linewidth}\raggedright
\textbf{Val F1}
\end{minipage} \\
\midrule\noalign{}
\endhead
\bottomrule\noalign{}
\endlastfoot
\textbf{Ensemble}

\textbf{(Fusion-DGCNN+XGBoost+PGLR)} & \textbf{0.87 ± 0.01} & \textbf{0.86 ± 0.01} & \textbf{0.86 ± 0.01} & \textbf{0.86 ± 0.01} \\
Fusion-PointBERTStyle & 0.77 ± 0.08 & 0.78 ± 0.06 & 0.77 ± 0.08 & 0.77 ± 0.08 \\
MLP & 0.77 ± 0.11 & 0.80 ± 0.11 & 0.77 ± 0.11 & 0.77 ± 0.11 \\
Fusion-PointNet & 0.75 ± 0.10 & 0.77 ± 0.09 & 0.76 ± 0.10 & 0.75 ± 0.10 \\
XGBoost & 0.75 ± 0.11 & 0.78 ± 0.12 & 0.77 ± 0.11 & 0.76 ± 0.11 \\
Fusion-PointNet++ & 0.74 ± 0.09 & 0.75 ± 0.08 & 0.74 ± 0.08 & 0.74 ± 0.09 \\
Fusion-DGCNN & 0.74 ± 0.09 & 0.75 ± 0.08 & 0.74 ± 0.08 & 0.74 ± 0.09 \\
PGLR & 0.74 ± 0.07 & 0.77 ± 0.08 & 0.76 ± 0.07 & 0.74 ± 0.07 \\
LightGBM & 0.72 ± 0.13 & 0.75 ± 0.14 & 0.74 ± 0.13 & 0.73 ± 0.13 \\
Pure\_PointNet & 0.32 ± 0.03 & 0.11 ± 0.01 & 0.33 ± 0.00 & 0.16 ± 0.01 \\
\end{longtable}

Table 7. Comparison of 2D+3D model performance in test set, mean ± SD across 5 folds (n=5). 95\% CI computed via the t-distribution.

\begin{longtable}[]{@{}
  >{\raggedright\arraybackslash}p{(\columnwidth - 10\tabcolsep) * \real{0.2305}}
  >{\raggedright\arraybackslash}p{(\columnwidth - 10\tabcolsep) * \real{0.1733}}
  >{\raggedright\arraybackslash}p{(\columnwidth - 10\tabcolsep) * \real{0.1636}}
  >{\raggedright\arraybackslash}p{(\columnwidth - 10\tabcolsep) * \real{0.1444}}
  >{\raggedright\arraybackslash}p{(\columnwidth - 10\tabcolsep) * \real{0.1444}}
  >{\raggedright\arraybackslash}p{(\columnwidth - 10\tabcolsep) * \real{0.1439}}@{}}
\toprule\noalign{}
\begin{minipage}[b]{\linewidth}\raggedright
\textbf{Model}
\end{minipage} & \begin{minipage}[b]{\linewidth}\raggedright
\textbf{Test Acc.}
\end{minipage} & \begin{minipage}[b]{\linewidth}\raggedright
\textbf{Test 95\% CI}
\end{minipage} & \begin{minipage}[b]{\linewidth}\raggedright
\textbf{Test Prec.}
\end{minipage} & \begin{minipage}[b]{\linewidth}\raggedright
\textbf{Test Rec.}
\end{minipage} & \begin{minipage}[b]{\linewidth}\raggedright
\textbf{Test F1}
\end{minipage} \\
\midrule\noalign{}
\endhead
\bottomrule\noalign{}
\endlastfoot
\textbf{Ensemble (FusionDGCNN+XGBoost+PGLR)} & \textbf{0.90 ± 0.03} & \textbf{{[}0.85, 0.93{]}} & \textbf{0.89 ± 0.04} & \textbf{0.88 ± 0.03} & \textbf{0.88 ± 0.03} \\
LightGBM & 0.84 ± 0.07 & {[}0.75, 0.93{]} & 0.86 ± 0.06 & 0.84 ± 0.07 & 0.84 ± 0.07 \\
XGBoost & 0.83 ± 0.04 & {[}0.78, 0.88{]} & 0.84 ± 0.04 & 0.82 ± 0.05 & 0.83 ± 0.04 \\
Fusion-DGCNN & 0.80 ± 0.08 & {[}0.71, 0.90{]} & 0.80 ± 0.10 & 0.79 ± 0.08 & 0.79 ± 0.09 \\
PGLR & 0.80 ± 0.09 & {[}0.69, 0.92{]} & 0.83 ± 0.08 & 0.80 ± 0.10 & 0.80 ± 0.09 \\
Fusion-PointNet++ & 0.78 ± 0.11 & {[}0.64, 0.93{]} & 0.79 ± 0.12 & 0.78 ± 0.11 & 0.78 ± 0.12 \\
Fusion-PointNet & 0.76 ± 0.08 & {[}0.67, 0.86{]} & 0.77 ± 0.08 & 0.75 ± 0.07 & 0.75 ± 0.08 \\
Fusion-PointBERTStyle & 0.75 ± 0.07 & {[}0.66, 0.83{]} & 0.74 ± 0.08 & 0.73 ± 0.07 & 0.73 ± 0.07 \\
MLP & 0.71 ± 0.04 & {[}0.66, 0.75{]} & 0.70 ± 0.05 & 0.69 ± 0.04 & 0.69 ± 0.04 \\
Pure\_PointNet & 0.32 ± 0.02 & {[}0.30, 0.35{]} & 0.11 ± 0.01 & 0.33 ± 0.00 & 0.16 ± 0.01 \\
\end{longtable}

The Fusion-DGCNN + XGBoost + PGLR ensemble achieved the strongest overall performance, with a validation accuracy of 0.87 ± 0.01 and test accuracy of 0.89 ± 0.03, demonstrating both high predictive performance and low cross-fold variability as shown in Tables 6 and 7. Among the individual models, LightGBM showed the strongest test generalization, achieving 0.84 ± 0.07 accuracy despite a lower validation accuracy of 0.72 ± 0.13, while XGBoost similarly improved from 0.75 ± 0.11 validation accuracy to 0.83 ± 0.04 on the test folds. In contrast, several point-cloud models exhibited lower and more variable performance, suggesting greater sensitivity to the limited sample size and fold composition. The superior and more consistent performance of the ensemble likely resulted from combining complementary information captured by Fusion-DGCNN, XGBoost, and PGLR, allowing errors from individual models to be partially compensated while integrating learned 3D representations with image-derived and physics-informed information.

Table 8. Statistical significance of the Ensemble combination\textquotesingle s test accuracy vs. each individual 2D+3D model (paired t-test, n=5 folds).

\begin{longtable}[]{@{}
  >{\raggedright\arraybackslash}p{(\columnwidth - 4\tabcolsep) * \real{0.2786}}
  >{\raggedright\arraybackslash}p{(\columnwidth - 4\tabcolsep) * \real{0.3881}}
  >{\raggedright\arraybackslash}p{(\columnwidth - 4\tabcolsep) * \real{0.3333}}@{}}
\toprule\noalign{}
\begin{minipage}[b]{\linewidth}\raggedright
\textbf{Model}
\end{minipage} & \begin{minipage}[b]{\linewidth}\raggedright
\textbf{Mean diff. vs. Ensemble (pp)}
\end{minipage} & \begin{minipage}[b]{\linewidth}\raggedright
\textbf{Paired t-test p-value}
\end{minipage} \\
\midrule\noalign{}
\endhead
\bottomrule\noalign{}
\endlastfoot
Fusion-PointNet & +13.57 & 0.015 \\
Pure\_PointNet & +57.67 & \textless0.001 \\
PointNet++ & +11.52 & 0.063 \\
Fusion-DGCNN & +9.62 & 0.049 \\
PointBERT & +15.43 & 0.004 \\
XGBoost & +6.62 & 0.008 \\
LightGBM & +5.71 & 0.185 \\
MLP & +19.48 & 0.003 \\
PGLR & +9.62 & 0.100 \\
\end{longtable}

\includegraphics[width=4.67766in,height=2.32911in]{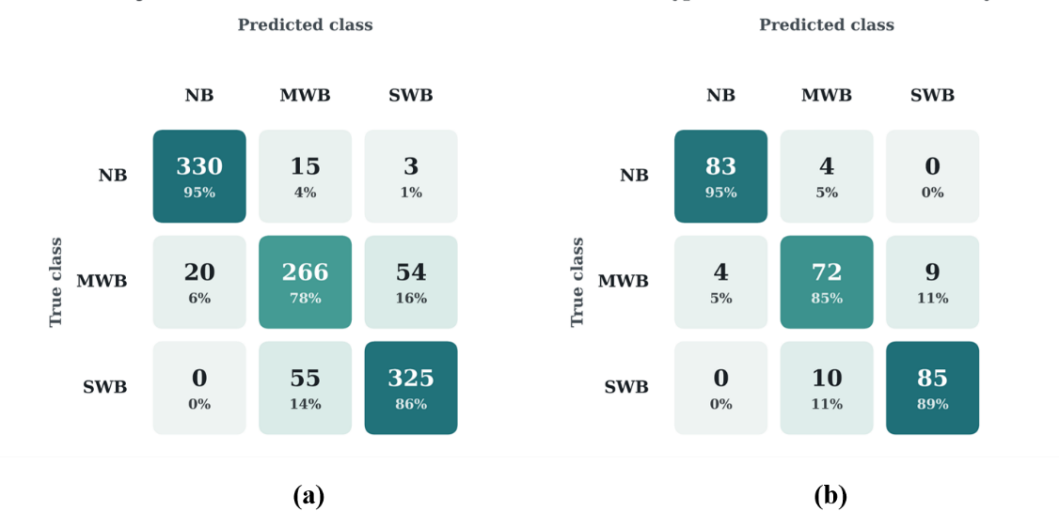}

Fig. 11 Confusion matrix of Ensemble model (Fusion-DGCNN + XGBoost + PGLR) (a) validation set and (b) test set

The Fusion-DGCNN + XGBoost + PGLR ensemble showed balanced discrimination across the three WB severity classes, with test recalls of 95\% for NB, 85\% for MWB, and 89\% for SWB. Only four NB samples were incorrectly assigned to MWB, indicating a clear separation between healthy and myopathy-affected fillets. Classification errors were concentrated primarily at the MWB-SWB boundary, where 9 MWB samples were predicted as SWB and 10 SWB samples as MWB as shown in Fig. 11. This pattern suggests that the ensemble effectively distinguishes normal tissue while retaining improved sensitivity to different levels of WB severity. Its strong performance likely results from the complementary information provided by the constituent models, integrating 3D point-cloud representations, image-derived characteristics, and physics-informed features into the final prediction.

\hypertarget{ablation-study}{%
\subsection{\texorpdfstring{Ablation Study }{Ablation Study }}\label{ablation-study}}

Table 9. Marginal contribution of each model within the final ensemble combination per-fold test accuracy.

\begin{longtable}[]{@{}
  >{\raggedright\arraybackslash}p{(\columnwidth - 8\tabcolsep) * \real{0.0861}}
  >{\raggedright\arraybackslash}p{(\columnwidth - 8\tabcolsep) * \real{0.2503}}
  >{\raggedright\arraybackslash}p{(\columnwidth - 8\tabcolsep) * \real{0.2636}}
  >{\raggedright\arraybackslash}p{(\columnwidth - 8\tabcolsep) * \real{0.2000}}
  >{\raggedright\arraybackslash}p{(\columnwidth - 8\tabcolsep) * \real{0.2000}}@{}}
\toprule\noalign{}
\begin{minipage}[b]{\linewidth}\raggedright
\textbf{Fold}
\end{minipage} & \begin{minipage}[b]{\linewidth}\raggedright
\textbf{Full combo Test acc.}
\end{minipage} & \begin{minipage}[b]{\linewidth}\raggedright
\textbf{w/o Fusion-DGCNN}
\end{minipage} & \begin{minipage}[b]{\linewidth}\raggedright
\textbf{w/o XGBoost}
\end{minipage} & \begin{minipage}[b]{\linewidth}\raggedright
\textbf{w/o PGLR}
\end{minipage} \\
\midrule\noalign{}
\endhead
\bottomrule\noalign{}
\endlastfoot
1 & 0.86 & −1.26 & +3.79 & +3.79 \\
2 & 0.88 & +6.56 & +5.26 & +7.88 \\
3 & 0.90 & +3.91 & +5.22 & +3.91 \\
4 & 0.92 & +6.48 & +3.89 & +3.89 \\
5 & 0.94 & −1.36 & +1.36 & +2.73 \\
\end{longtable}

Table 10. Mean ± SD marginal contribution of each model, across the 5 folds, ranked highest to lowest.

\begin{longtable}[]{@{}
  >{\raggedright\arraybackslash}p{(\columnwidth - 4\tabcolsep) * \real{0.2497}}
  >{\raggedright\arraybackslash}p{(\columnwidth - 4\tabcolsep) * \real{0.4170}}
  >{\raggedright\arraybackslash}p{(\columnwidth - 4\tabcolsep) * \real{0.3333}}@{}}
\toprule\noalign{}
\begin{minipage}[b]{\linewidth}\raggedright
\textbf{Model}
\end{minipage} & \begin{minipage}[b]{\linewidth}\raggedright
\textbf{Mean marginal contribution (pp)}
\end{minipage} & \begin{minipage}[b]{\linewidth}\raggedright
\textbf{Std (pp)}
\end{minipage} \\
\midrule\noalign{}
\endhead
\bottomrule\noalign{}
\endlastfoot
PGLR & +4.44 & 1.99 \\
XGBoost & +3.90 & 1.58 \\
Fusion-DGCNN & +2.87 & 3.96 \\
\end{longtable}

As shown in Table 7 PGLR a 5-physics based feature logistic regression, the simplest and individually weakest of the three models in the table, contributes the most on average (+4.44 pp) and is positive in every fold. XGBoost is the most consistent contributor, with the lowest standard deviation, 1.58 pp, positive in every fold, as shown in Tables 9 and 10. Fusion-DGCNN, despite being the highest-capacity model of the three, contributes the least on average (+2.87 pp) and is the least reliable. This mirrors the pattern already established across all 511 combinations, the point-cloud deep-learning models are the least reliable contributors regardless of overall accuracy level. The combination\textquotesingle s stability is driven more by its two feature-based members, XGBoost and PGLR, than by its deep-learning member, Fusion-DGCNN.

Table 11. Mean marginal contribution of each model to ensemble accuracy (pp), averaged across 5 folds.

\begin{longtable}[]{@{}
  >{\raggedright\arraybackslash}p{(\columnwidth - 4\tabcolsep) * \real{0.3334}}
  >{\raggedright\arraybackslash}p{(\columnwidth - 4\tabcolsep) * \real{0.3334}}
  >{\raggedright\arraybackslash}p{(\columnwidth - 4\tabcolsep) * \real{0.3333}}@{}}
\toprule\noalign{}
\begin{minipage}[b]{\linewidth}\raggedright
\textbf{Model}
\end{minipage} & \begin{minipage}[b]{\linewidth}\raggedright
\textbf{Mean contribution (pp)}
\end{minipage} & \begin{minipage}[b]{\linewidth}\raggedright
\textbf{2D+3D / 2D-only}
\end{minipage} \\
\midrule\noalign{}
\endhead
\bottomrule\noalign{}
\endlastfoot
LightGBM & +2.78 ± 0.64 & 2D+3D \\
PGLR & +2.36 ± 0.64 & 2D+3D \\
XGBoost & +2.08 ± 0.63 & 2D+3D \\
MLP & +1.31 ± 0.90 & 2D+3D \\
Fusion-PointBERTStyle & +0.07 ± 0.74 & 2D+3D \\
Fusion-DGCNN & +0.00 ± 0.57 & 2D+3D \\
Pure\_PointNet & -0.17 ± 0.06 & 2D+3D \\
Fusion-PointNet & -0.63 ± 0.77 & 2D+3D \\
Fusion-PointNet++ & -0.88 ± 0.62 & 2D+3D \\
PGLR & +3.32 ± 2.04 & 2D-only \\
XGBoost & +1.42 ± 0.89 & 2D-only \\
LightGBM & +1.35 ± 1.78 & 2D-only \\
MLP & -2.31 ± 1.94 & 2D-only \\
\end{longtable}

Table 12. Top 10 of all 511 2D+3D combinations, ranked by 5-fold mean out-of-fold accuracy and marginal contribution (mean ± SD across 5 folds) of each model.

\begin{longtable}[]{@{}
  >{\raggedright\arraybackslash}p{(\columnwidth - 8\tabcolsep) * \real{0.0831}}
  >{\raggedright\arraybackslash}p{(\columnwidth - 8\tabcolsep) * \real{0.3206}}
  >{\raggedright\arraybackslash}p{(\columnwidth - 8\tabcolsep) * \real{0.0674}}
  >{\raggedright\arraybackslash}p{(\columnwidth - 8\tabcolsep) * \real{0.1540}}
  >{\raggedright\arraybackslash}p{(\columnwidth - 8\tabcolsep) * \real{0.3749}}@{}}
\toprule\noalign{}
\begin{minipage}[b]{\linewidth}\raggedright
\textbf{Rank}
\end{minipage} & \begin{minipage}[b]{\linewidth}\raggedright
\textbf{Combination}
\end{minipage} & \begin{minipage}[b]{\linewidth}\raggedright
\textbf{Size}
\end{minipage} & \begin{minipage}[b]{\linewidth}\raggedright
\textbf{Mean OOF accuracy}
\end{minipage} & \begin{minipage}[b]{\linewidth}\raggedright
\textbf{Marginal contribution within this combination (pp)}
\end{minipage} \\
\midrule\noalign{}
\endhead
\bottomrule\noalign{}
\endlastfoot
\textbf{1} & Fusion-DGCNN + XGBoost + PGLR & 3 & 90.00 ± 0.03 & Fusion-DGCNN +2.87±3.96; XGBoost +3.90±1.58; PGLR +4.44±1.99 \\
2 & Fusion-PointBERTStyle + LightGBM + PGLR & 3 & 84.31 ± 1.61 & Fusion-PointBERTStyle +2.21±1.34; LightGBM +5.88±1.35; PGLR +3.43±1.03 \\
3 & Pure\_PointNet + Fusion-DGCNN + XGBoost + PGLR & 4 & 84.31 ± 1.41 & Pure\_PointNet +0.00±0.00; Fusion-DGCNN +2.69±3.72; XGBoost +3.68±1.51; PGLR +3.93±2.69 \\
4 & Pure\_PointNet + Fusion-DGCNN + XGBoost + LightGBM + PGLR & 5 & 84.31 ± 1.69 & Pure\_PointNet +0.49±0.68; Fusion-DGCNN +1.71±2.04; XGBoost +0.73±1.39; LightGBM −0.00±1.22; PGLR +2.45±0.86 \\
5 & Pure\_PointNet + Fusion-PointBERTStyle + LightGBM + PGLR & 4 & 84.07 ± 1.73 & Pure\_PointNet −0.24±0.55; Fusion-PointBERTStyle +1.47±1.59; LightGBM +5.64±1.65; PGLR +3.43±1.02 \\
6 & Fusion-PointNet + Pure\_PointNet + LightGBM + PGLR & 4 & 84.07 ± 1.97 & Fusion-PointNet +1.47±1.02; Pure\_PointNet +0.74±1.10; LightGBM +6.39±4.57; PGLR +3.69±1.96 \\
7 & Fusion-PointBERTStyle + XGBoost + LightGBM + PGLR & 4 & 84.07 ± 1.28 & Fusion-PointBERTStyle +1.47±1.61; XGBoost −0.25±1.02; LightGBM +1.23±3.15; PGLR +1.72±1.41 \\
8 & Pure\_PointNet + Fusion-PointBERTStyle + XGBoost + LightGBM + PGLR & 5 & 84.07 ± 1.28 & Pure\_PointNet +0.00±0.00; Fusion-PointBERTStyle +1.47±1.61; XGBoost −0.00±1.50; LightGBM +0.98±2.68; PGLR +1.96±1.65 \\
9 & Pure\_PointNet + Fusion-DGCNN + Fusion-PointBERTStyle + LightGBM + MLP + PGLR & 6 & 83.83 ± 1.54 & Pure\_PointNet +0.49±0.67; Fusion-DGCNN +0.74±1.10; Fusion-PointBERTStyle +1.00±3.18; LightGBM +2.45±1.74; MLP +1.72±1.41; PGLR +2.70±1.83 \\
10 & Fusion-PointNet + Fusion-PointBERTStyle + LightGBM + MLP + PGLR & 5 & 83.82 ± 2.51 & Fusion-PointNet +0.49±2.06; Fusion-PointBERTStyle +1.23±2.75; LightGBM +2.70±2.37; MLP +2.46±2.90; PGLR +1.72±1.86 \\
\end{longtable}

The final recommended combination, Fusion-DGCNN + XGBoost + PGLR, ranked first with a 5-fold mean OOF accuracy of 90.00\% ± 0.03 (table 12), substantially exceeding the next-best combinations at 84.31\%. PGLR provided the largest contribution and remained positive across all top-10 combinations, indicating a particularly consistent complementary role. In contrast, adding Pure PointNet did not improve the winning ensemble, while XGBoost became redundant or slightly detrimental in some combinations containing models with overlapping information. These findings demonstrate that model complementarity, rather than ensemble size, was critical to performance.

The winning ensemble effectively combines complementary representations: Fusion-DGCNN captures learned multimodal 2D+3D information, XGBoost models nonlinear relationships among structured features, and PGLR incorporates physics-informed relationships. Although LightGBM contributed strongly in several alternative combinations, it was absent from the highest-performing ensemble, further showing that strong individual contribution does not necessarily translate to the optimal model combination. Overall, the compact three-model ensemble achieved the highest predictive performance while limiting redundancy and computational complexity.

Table 13. Standalone PurePointNet test accuracy vs. training-set size, mean ± SD across 5 folds.

\begin{longtable}[]{@{}
  >{\raggedright\arraybackslash}p{(\columnwidth - 6\tabcolsep) * \real{0.2501}}
  >{\raggedright\arraybackslash}p{(\columnwidth - 6\tabcolsep) * \real{0.2499}}
  >{\raggedright\arraybackslash}p{(\columnwidth - 6\tabcolsep) * \real{0.2501}}
  >{\raggedright\arraybackslash}p{(\columnwidth - 6\tabcolsep) * \real{0.2499}}@{}}
\toprule\noalign{}
\begin{minipage}[b]{\linewidth}\raggedright
\textbf{Training fraction}
\end{minipage} & \begin{minipage}[b]{\linewidth}\raggedright
\textbf{Median n (train)}
\end{minipage} & \begin{minipage}[b]{\linewidth}\raggedright
\textbf{Mean test accuracy}
\end{minipage} & \begin{minipage}[b]{\linewidth}\raggedright
\textbf{95\% CI}
\end{minipage} \\
\midrule\noalign{}
\endhead
\bottomrule\noalign{}
\endlastfoot
25\% & 16 & 0.30 ± 0.03 & {[}0.27, 0.34{]} \\
50\% & 32 & 0.30 ± 0.02 & {[}0.28, 0.33{]} \\
75\% & 48 & 0.29 ± 0.01 & {[}0.28, 0.30{]} \\
100\% & 65 & 0.29 ± 0.01 & {[}0.28, 0.30{]} \\
\end{longtable}

The PurePointNet training-size ablation showed consistently poor performance across all training fractions, with mean test accuracy remaining between 0.29 and 0.30 as the number of training samples increased approximately fourfold, from 45 samples at 25\% to 181 samples at 100\% as shown in Table 13. Increasing the available training data produced no measurable improvement, with overlapping 95\% confidence intervals across all conditions. This suggests that PurePointNet\textquotesingle s near-chance performance is unlikely to be explained primarily by limited training-set size. Instead, the results indicate an architectural limitation in extracting sufficiently discriminative WB severity information from raw point coordinates alone at the evaluated sample scale. This is also consistent with the ensemble analysis, where PurePointNet provided essentially no additional benefit to the final Fusion-DGCNN + XGBoost + PGLR combination.

\hypertarget{grad-cam-visualization}{%
\subsection{Grad-CAM Visualization}\label{grad-cam-visualization}}

Grad-CAM was computed for all 5 point-cloud architectures (Fusion-PointNet, Pure\_PointNet, Fusion-PointNet++, Fusion-DGCNN, Fusion-PointBERTStyle) on 3 held-out test samples, one per class (NB\_11, MW\_01, SW\_02, all from fold 5), each scored against all 3 class hypotheses 45 renders in total. Every model now has full-cloud coverage, Fusion-PointNet++ and Fusion-PointBERTStyle originally used a coarse patch-based grouping that only covered part of the cloud and were rebuilt to use nearest-center (Voronoi) assignment so every point in every cloud is traceable back to an importance value, matching the full coverage that Fusion-PointNet, Pure\_PointNet, and Fusion-DGCNN already had.

\includegraphics[width=4.49982in,height=2.84494in]{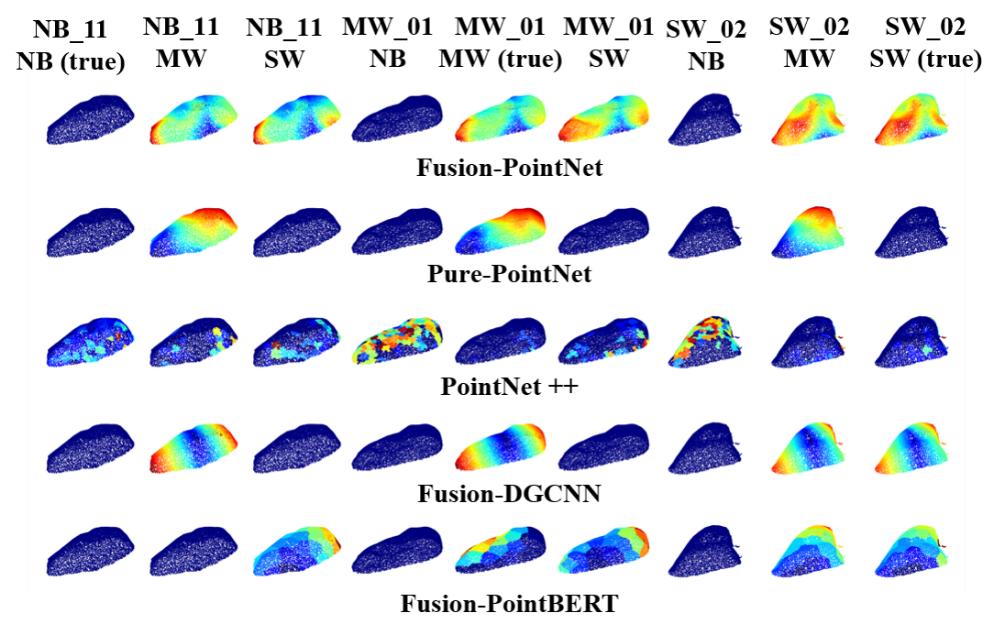}

Figure 4. Grad-CAM comparison across all 5 point-cloud models. Rows = model, grouped by sample (3 blocks of 3 columns); true headers mark the ground-truth class for that sample.

One interpretive caveat applies to four of these five panels before reading anything into their shape. Grad-CAM\textquotesingle s activations are captured purely from each model\textquotesingle s point-cloud branch but for Fusion-PointNet, Fusion-PointNet++, Fusion-DGCNN, and Fusion-PointBERTStyle, that branch\textquotesingle s weights were never trained in isolation. All four are optimized end-to-end with a single shared loss computed from the fused (2D+3D) output, and one backward pass per training step sends gradients through the entire network at once the point-cloud layers included so each of these four point-cloud branches was shaped throughout training by what the 2D branch was simultaneously contributing to the same loss, even though the point-cloud branch itself never sees the 2D features directly. Pure\_PointNet is the only one of the five that was never trained alongside a 2D branch at all, so it is the only panel in this figure showing what a point-cloud-only decision process actually looks like in isolation; the other four show a point-cloud branch that had a 2D partner to lean on throughout training.

Four of the five models (Fusion-PointNet, Pure\_PointNet, Fusion-DGCNN, Fusion-PointBERTStyle) show a single smooth gradient per panel, running from the fillet\textquotesingle s thicker base to its tapered tip consistent with the attenuation signal identified as the top SHAP feature in Section 3.2.2. Fusion-PointNet++ is a clear outlier, its saliency is scattered and patchy small, disconnected high-importance clusters rather than one coherent region in nearly every panel, including its own ground-truth-class panels.

Each model\textquotesingle s visual pattern traces directly to where in its architecture Grad-CAM had to be hooked, not to an arbitrary stylistic difference. Pure\_PointNet, Fusion-PointNet, and Fusion-DGCNN all still have a genuine per-point feature map immediately before their final global max-pool (conv5), so Grad-CAM there produces one continuous importance value per point the finest resolution the data allows. PointNet++ has no such layer: it downsamples the cloud twice before pooling (first to 512 local cluster centers, then to 128 super-cluster centers), so Grad-CAM can only be computed on those 128 super-clusters, and every point in the cloud inherits its value by tracing back through both downsampling stages to whichever of the 128 centers it was assigned to. The result is a piecewise-constant map with only 128 distinct values covering the entire cloud, and because the two downsampling stages were learned independently of any global spatial trend, neighboring points that happen to fall into different clusters can land on very different values producing exactly the scattered, disconnected patches visible in the figure, not visualization noise. Fusion-PointNet++\textquotesingle s patchiness is therefore a direct, mechanical consequence of two-stage local set-abstraction, the same architectural property already flagged as this model\textquotesingle s key structural difference from the other four.

Fusion-PointBERTStyle is architecturally closer to PointNet++ than it looks, it also has no per-point conv5, and Grad-CAM is likewise computed on a coarse set of regions 32 patches, assigned by nearest-center partitioning of the whole cloud fewer regions than PointNet++\textquotesingle s 128, yet its panels still read as one smooth gradient rather than as 32 visible blocks. Two architectural differences explain why. First, PointBERTStyle\textquotesingle s 32 patch centers are chosen directly from the raw cloud in a single step (farthest-point sampling), so each patch is one large, spatially compact region of the fillet; PointNet++\textquotesingle s 128 regions are the end product of two independently learned downsampling steps chained together, and errors or irregularities introduced at the first stage propagate into the second, breaking up what would otherwise be spatially coherent regions. Second, the true signal being approximated here a gradient along the fillet\textquotesingle s long axis, from thick base to tapered tip is a low-frequency spatial trend, and 32 large, contiguous regions laid end-to-end along that axis can still trace out a visibly smooth ramp; only a genuinely irregular or high-frequency importance pattern would expose a 32-patch quantization as visibly blocky the way PointNet++\textquotesingle s does. Fusion-DGCNN, meanwhile, sits closest to Pure\_PointNet and Fusion-PointNet in mechanism (genuine per-point resolution, no patching) but is the only one of the three whose per-point feature is built by explicitly aggregating each point\textquotesingle s 20 nearest neighbors at every layer, with the neighbor graph rebuilt in feature space rather than fixed to the raw 3D coordinates. That gives Fusion-DGCNN\textquotesingle s per-point features access to a wider effective neighborhood than Pure\_PointNet or Fusion-PointNet\textquotesingle s purely independent per-point processing, without sacrificing the point-level resolution that keeps its Grad-CAM smooth.

These architectural explanations line up with the real quantitative results. Fusion-PointNet++ exhibited patchy Grad-CAM activation, consistent with its relatively unstable test performance (0.78 ± 0.11) and poor ensemble contribution. Fusion-DGCNN and Fusion-PointBERTStyle, whose Grad-CAM is smooth and coherent, are correspondingly the two least-negative point-cloud contributors to the ensemble (+0.00 ± 0.53 and +0.07 ± 0.69 pp respectively) the closest point-cloud model comes to being a genuinely useful ensemble member. Fusion-PointNet is the one exception worth flagging rather than glossing over, its Grad-CAM is just as smooth and coherent as Fusion-DGCNN\textquotesingle s and Fusion-PointBERTStyle\textquotesingle s, yet it is the second-worst ensemble contributor (−0.59 ± 0.72 pp). A smooth Grad-CAM indicates a consistent learned pattern, but not necessarily complementary information. Fusion-PointNet appears to capture the same base-to-tip density pattern already represented by the attenuation feature, making its contribution potentially redundant. Joint training with a strong 2D branch may further encourage the point-cloud branch to rely on this readily available signal rather than learning distinct geometric cues.

A second, more general pattern across all five models, the MW and SW class-hypothesis panels consistently activate on overlapping or comparably strong regions for the same sample, while the NB hypothesis is visually distinct. This mirrors the confusion matrix in Fig. 11 exactly MW is confused almost entirely with SW, never with NB and the fact that five architecturally different models (three global-pooling variants, one local-graph model, one patch-based transformer) all converge on the same failure mode is stronger evidence that the MW/SW ambiguity reflects a genuine property of the geometry near that class boundary, rather than an artifact of any one architecture\textquotesingle s inductive bias.

A third, for a handful of (sample, class) panels, Grad-CAM returns a flat, uniform map with no visible gradient most often on a sample\textquotesingle s own ground-truth-class panel rather than an incorrect one. This is an expected property of the underlying method, not a rendering error. Grad-CAM applies a ReLU to the gradient-weighted activation map, keeping only positive-contributing evidence; if every point\textquotesingle s weighted activation for that class is non-positive, the entire map is zeroed out and normalization is skipped. A flat panel means no single region drove that class\textquotesingle s score upward under this linear-approximation method it does not mean the model had zero confidence in that class, only that whatever evidence it used was diffuse, bias-driven, or otherwise not captured by this attribution technique.

The results demonstrate a clear improvement from approximately 82\% accuracy using 2D transillumination features alone to 90\% with the multimodal ensemble, indicating that combining optical information with 3D geometric and physics-informed features provides complementary information for WB severity classification. The lower performance of the 2D-only approach likely reflects both the limited information available from transillumination and the relatively small dataset of 267 samples. Classification remained particularly challenging for MWB because it represents an intermediate and heterogeneous stage whose optical and structural characteristics overlap with both NB and SWB, while localized abnormalities may be diluted when features are calculated across the entire fillet. Subjectivity associated with palpation-based ground truth may further increase uncertainty for borderline cases. Although multimodal fusion substantially improved discrimination by incorporating geometric information unavailable from 2D imaging alone, some ambiguity between adjacent severity classes remained due to the biological continuum of WB. Further improvements will therefore require larger and more diverse datasets, objective reference measurements, and potentially continuous severity scoring.

Based on the literature review, as shown in Table 14, most existing single-modality approaches using traditional machine learning achieve 73-91\% accuracy for severity classification, while advanced deep learning and spectroscopy methods reach 93-100\% but often only for binary detection rather than multi-class severity grading. To the best of our knowledge, this is also the first study to apply transillumination imaging captured via a consumer mobile phone for woody breast classification; a small number of related studies have used optical imaging for this task notably light scattering imaging and structured-illumination reflectance imaging, but these rely on a fundamentally different acquisition geometry (light reflected back from the tissue surface rather than transmitted through the fillet) and neither has been implemented on a mobile, consumer-grade platform. The research gap in true multi-modal fusion, particularly combining image and 3D structural features, further positions our approach as novel in this domain.

Table 14. Comparison of Sensing Modalities and Machine Learning Methods used for WB Assessment

\begin{longtable}[]{@{}
  >{\raggedright\arraybackslash}p{(\columnwidth - 10\tabcolsep) * \real{0.2046}}
  >{\raggedright\arraybackslash}p{(\columnwidth - 10\tabcolsep) * \real{0.0729}}
  >{\raggedright\arraybackslash}p{(\columnwidth - 10\tabcolsep) * \real{0.1827}}
  >{\raggedright\arraybackslash}p{(\columnwidth - 10\tabcolsep) * \real{0.1333}}
  >{\raggedright\arraybackslash}p{(\columnwidth - 10\tabcolsep) * \real{0.1396}}
  >{\raggedright\arraybackslash}p{(\columnwidth - 10\tabcolsep) * \real{0.2669}}@{}}
\toprule\noalign{}
\begin{minipage}[b]{\linewidth}\raggedright
\textbf{Study}
\end{minipage} & \begin{minipage}[b]{\linewidth}\raggedright
\textbf{Year}
\end{minipage} & \begin{minipage}[b]{\linewidth}\raggedright
\textbf{Modality}
\end{minipage} & \begin{minipage}[b]{\linewidth}\raggedright
\textbf{Sample Size/Classes}
\end{minipage} & \begin{minipage}[b]{\linewidth}\raggedright
\textbf{Accuracy}
\end{minipage} & \begin{minipage}[b]{\linewidth}\raggedright
\textbf{Method}
\end{minipage} \\
\midrule\noalign{}
\endhead
\bottomrule\noalign{}
\endlastfoot
Liu et al. & 2025 & Hyperspectral & 150 / 3 & 77.8\% & LDA \\
Pallerla et al. & 2024 & Hyperspectral & 250 / 3 & 95\% & NAS-WD \\
Lu et al. & 2024 & BIA & 80 / 4 & 90\% & L-BFGS \\
Yoon et al. & 2022 & Bending Angle & 45 / 2 & 95\% & Sideview Imaging \\
Yang et al. & 2021 & EF & 216 / 3 & 93.3\% & Deep Learning

(AlexNet pre-trained) \\
Caldas-Cueva et al. & 2021 & RGB & 900 / 2 & 84\% & Logistic Reg. \\
Geronimo et al. & 2019 & RGB \& NIR & 80 / 2 & 91.8\% & SVM \\
\end{longtable}

The literature on woody breast (WB) detection clusters into three broad approaches, spectroscopic/hyperspectral methods, destructive laboratory assays, and external or mechanical imaging and a consistent pattern holds across all three severity classification (3-class or finer) underperforms binary detection by roughly 10-20 \% points, regardless of modality.

Near-infrared spectroscopy (NIRS) remains the industrial benchmark for binary WB detection. Wold et al. (2017, 2019) reported 96-100\% binary accuracy using the TOMRA QVision500 scanner, which processes \textasciitilde3.5 fillets per second by exploiting protein and water-binding signatures near the 980 nm absorption peak; three-class severity grading, however, has not been validated at that industrial speed. Hyperspectral studies consistently identify key wavelengths at 980 nm, 1440 nm, and 2000-2500 nm, corresponding to water absorption and molecular bonding changes associated with WB, and the Bound Water Index derived from these features correlates with NMR T2 relaxation measurements (r ≈ 0.78), reinforcing the biological basis of optical water-binding indicators. Against that binary benchmark, three-class severity work with the same modality shows the expected drop: Pallerla et al. (2024) achieved 80\% (SVM) and 73\% (MLP) for three-class WB classification via hyperspectral imaging, with their NAS-optimized Wide-Deep model (NAS-WD) reaching 95\% a 15-22 point improvement that demonstrates the value of architecture search over standard classifiers, but still on a single modality. Liu et al. (2025) reported the same binary-vs-severity gap directly, 91.67\% binary vs. only 77.78\% for three-class severity grading with hyperspectral imaging and LDA. Fusing modalities helps only partially, Geronimo et al. (2019) combined computer-vision features with NIR spectroscopy and reached 96.3\% higher than NIRS alone (91.8\%) but not higher than their best individual modality (97.5\%), indicating the two data sources were largely capturing overlapping, not complementary, information. Muñoz-Lapeira et al. (2025) compared NIR, fluorescence, and Raman spectroscopy directly, reporting 95-100\% binary accuracy across all three; fluorescence spectroscopy (350-580 nm) was uniquely sensitive to collagen crosslinking and adipose infiltration, biochemical markers tied directly to fibrosis severity, suggesting this kind of biochemical sensitivity could complement structural point-cloud features rather than duplicate them. Despite this accuracy ceiling, hyperspectral deployment remains costly and requires careful optical calibration, limiting practical adoption.

A second line of work trades non-destructiveness for higher accuracy on smaller, lab-scale samples. Yang et al. (2021) achieved 93.3\% for three-class WB severity using an expressible-fluid (EF) method, but it requires destructive filter-paper pressing to extract fluid, ruling out inline industrial use. López et al. (2022) applied GLCM texture descriptors to scanning electron microscopy images and demonstrated statistical separation between normal and severe WB, but their workflow is destructive, slow, and reported no quantitative classification accuracy. Bioelectrical impedance analysis (BIA) offers a non-optical alternative, Siddique et al. (2022) reported 66.67-85\% accuracy across severity classes using plate-based BIA with SVM, with moderate WB consistently the hardest class to detect. Lu et al. (2024) compared a partial least squares (PLS) model against an L-BFGS-optimized deep learning model on impedance features in a four-class setting: PLS reached higher training accuracy (91.70\% vs. 88.35\% for L-BFGS) including perfect 100\% accuracy for the severe-WB training class and for the normal/severe test classes individually but generalized worse overall, with only 80.00\% overall test accuracy versus L-BFGS\textquotesingle s 90.00\%. This reversal is itself informative, the model with the better training fit was not the model that generalized better, and L-BFGS\textquotesingle s 90\% test result lands in the same 85-90\% band seen across several other modalities, consistent with a shared performance ceiling driven by biological variability in the tissue rather than by any single algorithm\textquotesingle s limitations.

A third line of work targets industrial-line deployment using conveyor-integrated imaging systems, although these approaches are generally limited to binary classification.

Caldas-Cueva et al. (2021) achieved 84\% accuracy using simple logistic regression on geometric features (length, width, angle, area) extracted from RGB images of 544 broiler carcasses conceptually similar to shape-based descriptors, but without texture information, and limited to binary classification (normal/mild vs. moderate/severe merged). Yoon et al. (2022) developed a side-view imaging system measuring physical fillet bending, achieving 95\% accuracy at 100 feet/minute line speed via a normalized minimum distance measure (NMDM); the feature captures overall rigidity well but lacks the resolution to grade severity along a continuum, and the method is binary-only. Cai et al. (2025) combined structured-illumination reflectance imaging with 3D surface profilometry to reach 93\% accuracy, also binary-only, and while conceptually adjacent to point-cloud analysis this method characterizes surface topology rather than internal optical transmission properties.

Taken together, the literature shows that high accuracy is achievable for binary WB detection across nearly every modality, but reliable multi-class severity grading remains a harder, less-solved problem particularly under non-destructive, high-throughput constraints. The research gap in true multi-modal fusion, especially combining image-derived texture with 3D structural features, positions this study\textquotesingle s approach as novel, it combines transillumination texture features with Fusion-DGCNN-based point-cloud analysis, achieving competitive performance with the potential for practical severity classification in industrial poultry processing when integrated with a conveyor-based transillumination system and an industrial camera.

The proposed multimodal approach\textquotesingle s 90\% accuracy exceeds RGB-based morphometric methods (\textasciitilde84\%), traditional machine learning on spectral features (73-80\%), and bioelectrical impedance analysis (66-85\%), while approaching the performance of specialized deep learning systems (90-95\%) that require costly hyperspectral instrumentation or destructive sample preparation. This balance of high accuracy, non-destructive assessment, and portability positions the proposed method as a promising smartphone-based approach for WB severity classification, though further validation is needed under commercial processing conditions.

\hypertarget{future-works}{%
\section{Future Works}\label{future-works}}

The current study has several limitations that should be addressed to improve model robustness and support practical deployment. The dataset of 267 samples may not fully represent the phenotypic variability of Woody Breast (WB), particularly within the challenging MWB class. Future studies should therefore expand data collection across commercial processing facilities to incorporate variability associated with bird strain, age, diet, season, and processing conditions. In addition, reliance on manual palpation for ground-truth classification introduces potential subjectivity and inter-observer variability.

Although the present study evaluated multiple advanced point-cloud architectures, including DGCNN and an attention-based PointBERT-style model, further improvements could focus on more tightly integrated multimodal learning strategies. Cross-modal attention and intermediate feature-level fusion could enable direct interactions between transillumination-derived image information and 3D geometric representations, rather than relying primarily on the fusion of independently learned representations. Larger and more diverse datasets would also enable these more complex architectures to be investigated with a reduced risk of overfitting.

The smartphone-based framework was developed primarily to determine whether a consumer-grade mobile device can provide a feasible and portable platform for multimodal WB assessment. Although the acquisition approach could potentially be implemented on iPhone 12 and later models with the required imaging and depth-sensing capabilities, the present study was conducted and validated exclusively using an iPhone 16 Pro. Therefore, the reported performance should not be interpreted as validated across different iPhone generations. Differences in camera systems, image processing, and depth-sensing characteristics may influence the extracted 2D and 3D information. Future work should consequently include cross-device validation and, where necessary, device-specific calibration or model adaptation.

For practical implementation, particularly in small poultry processing plants, the computational requirements could be reduced through cloud-based inference. Images and point-cloud data acquired using the smartphone could be securely transferred to a cloud computing platform for model inference, with WB severity predictions and confidence scores returned to the mobile application. This approach could eliminate the need for individual facilities to purchase and maintain expensive local GPU workstations while allowing computational resources to be scaled according to demand. Collectively, these developments could provide a portable and potentially cost-effective framework for WB assessment while facilitating the collection of larger and more diverse datasets needed to improve model robustness and generalizability under commercial processing conditions.

\hypertarget{conclusion}{%
\section{Conclusion}\label{conclusion}}

In summary, this study demonstrates that integrating 2D transillumination-derived information, 3D point-cloud representations, and physics-informed features provides an effective approach for non-destructive three-class Woody Breast (WB) severity classification. Among the individual models, LightGBM achieved the highest test accuracy of 84\%, while the final Fusion-DGCNN + XGBoost + PGLR ensemble achieved the best overall performance at approximately 90\% accuracy. The ensemble achieved class recalls of 95\%, 85\%, and 89\% for NB, MWB, and SWB, respectively, with most errors occurring between MWB and SWB due to the overlapping characteristics of adjacent WB severity levels.

Evaluation of multiple 3D architectures further showed that model selection is critical for extracting meaningful information from fillet point clouds. While Pure PointNet remained near chance-level performance across different training-set sizes, DGCNN effectively captured local geometric relationships and became an important component of the best-performing fusion ensemble. Ensemble analysis further demonstrated that adding more models did not necessarily improve performance; instead, the compact combination of Fusion-DGCNN, XGBoost, and PGLR provided the highest accuracy, with each model contributing complementary information. Overall, the findings demonstrate the value of combining optical, geometric, and physics-informed information and support the feasibility of using a consumer-grade smartphone as a portable platform for non-destructive WB assessment. Future validation with larger and more diverse datasets, objective histological measurements, and different compatible smartphone generations will further establish the robustness and practical applicability of the proposed approach.

\hypertarget{acknowledgement}{%
\section{\texorpdfstring{Acknowledgement }{Acknowledgement }}\label{acknowledgement}}

The authors acknowledge the grant support by the University of Arkansas Experimental Station and the University of Arkansas College of Engineering,~USDA National Institute of Food and Agriculture~(No:~2023-70442-39232).

This research is supported by the Arkansas High Performance Computing Center which is funded through multiple National Science Foundation grants and the Arkansas Economic Development Commission.

\textbf{CRediT author statement}

Chaitanya Pallerla: Conceptualization, Methodology, Software, Investigation, Visualization, Writing - Original draft, Validation, Formal analysis.

Ramesh Bahadur Bist: Data curation

Yihong Feng: Data curation

Yang Tian: Data curation

Shiva Paudel: Software

Casey M. Owens: Resources.

Cengiz Koparan: Writing - Review \& Editing

Jeyamkondan Subbiah: Writing - Review \& Editing, Funding acquisition.

Dongyi Wang: Resources, Writing - Review \& Editing, Supervision, Project administration, Funding acquisition.

\hypertarget{references}{%
\section{References}\label{references}}

\begin{enumerate}
\def\labelenumi{\arabic{enumi}.}
\item
  Baldi, G., Soglia, F., \& Petracci, M. (2021). Spaghetti meat abnormality in broilers: Current understanding and future research directions. \emph{Frontiers in Physiology}, 12, 684497. \url{https://doi.org/10.3389/fphys.2021.684497}
\item
  Bhatt, M., Ayyalasomayajula, K.R., \& Yalavarthy, P.K. (2016). Generalized Beer-Lambert model for near-infrared light propagation in thick biological tissues. \emph{Journal of Biomedical Optics, 21(7), 076012.} \url{https://doi.org/10.1117/1.JBO.21.7.076012}
\item
  Breiman, L. (2001). Random forests. \emph{Machine Learning, 45}(1), 5-32. \url{https://doi.org/10.1023/A:1010933404324}
\item
  Cai, J., \& Lu, Y. (2024). Detection of Woody Breast Condition in Broiler Breast Fillets Using Light Scattering Imaging. \emph{Proceedings of SPIE, 13060}, 1306002. \url{https://doi.org/10.1117/12.3013464}
\item
  Cai, J., \& Lu, Y. (2025). Assessment of woody breast in broiler breast fillets using structured-illumination reflectance imaging coupled with surface profilometry. \emph{Journal of Food Engineering}, \emph{391}, Article 112459. \url{https://doi.org/10.1016/j.jfoodeng.2024.112459}
\item
  Caldas-Cueva, J. P., \& Owens, C. M. (2020). A review on the woody breast condition, detection methods, and product utilization in the contemporary poultry industry. \emph{Journal of Animal Science, 98}(7), skaa207. \url{https://doi.org/10.1093/jas/skaa207}
\item
  Caldas-Cueva, J. P., Mauromoustakos, A., Sun, X., \& Owens, C. M. (2021). Detection of woody breast condition in commercial broiler carcasses using image analysis. \emph{Poultry Science, 100}(3), 100977. \url{https://doi.org/10.1016/j.psj.2020.12.074}
\item
  Caldas-Cueva, J. P., Mauromoustakos, A., Sun, X., \& Owens, C. M. (2021). Use of image analysis to identify woody breast characteristics in 8-week-old broiler carcasses. \emph{Poultry Science, 100}(4), 100890. \url{https://doi.org/10.1016/j.psj.2020.12.003}
\item
  Canny, J. (1986). A computational approach to edge detection. \emph{IEEE Transactions on Pattern Analysis and Machine Intelligence}, 8(6), 679-698. \url{https://doi.org/10.1109/TPAMI.1986.4767851}
\item
  Chawla, N. V., Bowyer, K. W., Hall, L. O., \& Kegelmeyer, W. P. (2002). SMOTE: Synthetic minority over-sampling technique. \emph{Journal of Artificial Intelligence Research, 16}, 321-357. \url{https://doi.org/10.1613/jair.953}
\item
  Chen, T., \& Guestrin, C. (2016). XGBoost: A scalable tree boosting system. In \emph{Proceedings of the 22nd ACM SIGKDD International Conference on Knowledge Discovery and Data Mining} (pp. 785-794). ACM. \url{https://doi.org/10.1145/2939672.2939785}
\item
  Dalal, N., \& Triggs, B. (2005). Histograms of oriented gradients for human detection. In \emph{2005 IEEE Computer Society Conference on Computer Vision and Pattern Recognition (CVPR\textquotesingle05)} (Vol. 1, pp. 886-893). IEEE. \url{https://doi.org/10.1109/CVPR.2005.177}
\item
  DGCNN (Dynamic Graph CNN): Wang, Y., Sun, Y., Liu, Z., Sarma, S. E., Bronstein, M. M., \& Solomon, J. M. (2019). Dynamic Graph CNN for learning on point clouds. \emph{ACM Transactions on Graphics (TOG)}, \emph{38}(5), Article 146. \url{https://doi.org/10.1145/3326362}
\item
  Dietterich, T. G. (2000). Ensemble methods in machine learning. In \emph{Multiple Classifier Systems} (pp. 1-15). Springer. \url{https://doi.org/10.1007/3-540-45014-9_1}
\item
  Elsken, T., Metzen, J. H., \& Hutter, F. (2017). Simple and efficient architecture search for convolutional neural networks. \emph{arXiv:1711.04528}. \url{https://arxiv.org/abs/1711.04528}
\item
  Fischler, M. A., \& Bolles, R. C. (1981). Random sample consensus: A paradigm for model fitting with applications to image analysis and automated cartography. \emph{Communications of the ACM}\textbf{, 24}(6), 381-395. \url{https://doi.org/10.1145/358669.358692}
\item
  Geronimo, B. C., Mastelini, S. M., Carvalho, R. H., Barbon Júnior, S., Barbin, D. F., Shimokomaki, M., \& Ida, E. I. (2019). Computer vision system and near-infrared spectroscopy for identification and classification of chicken with wooden breast, and physicochemical and technological characterization. \emph{Infrared Physics \& Technology, 96}, 303-310. \url{https://doi.org/10.1016/j.infrared.2018.11.036}
\item
  Goodfellow, I., Bengio, Y., \& Courville, A. (2016). \emph{Deep learning}. MIT Press. \url{http://www.deeplearningbook.org}
\item
  Gulrajani, I., Ahmed, F., Arjovsky, M., Dumoulin, V., \& Courville, A. C. (2017). Improved training of Wasserstein GANs. \emph{Advances in Neural Information Processing Systems, 30}, 5767--5777.
\item
  Hadjidemetriou, E., Grossberg, M. D., \& Nayar, S. K. (2004). Multiresolution histograms and their use for recognition. \emph{IEEE Transactions on Pattern Analysis and Machine Intelligence, 26}(7), 831-847. \url{https://doi.org/10.1109/TPAMI.2004.32}
\item
  Han, H., Wang, W. Y., \& Mao, B. H. (2005). Borderline-SMOTE: A new over-sampling method in imbalanced data sets learning. In D. S. Huang, X. P. Zhang, \& G. B. Huang (Eds.), Advances in Intelligent Computing (Lecture Notes in Computer Science, Vol. 3644, pp. 878--887). \url{https://doi.org/10.1007/11538059_91}
\item
  Haralick, R. M., Shanmugam, K., \& Dinstein, I. (1973). Textural features for image classification. \emph{IEEE Transactions on Systems, Man, and Cybernetics}, SMC-3(6), 610-621. \url{https://doi.org/10.1109/TSMC.1973.4309314}
\item
  He, H., Bai, Y., Garcia, E. A., \& Li, S. (2008). ADASYN: Adaptive synthetic sampling approach for imbalanced learning. In \emph{2008 IEEE International Joint Conference on Neural Networks (IEEE World Congress on Computational Intelligence)} (pp. 1322--1328). https://doi.org/10.1109/IJCNN.2008.4633969
\item
  Hu, J., Shen, L., \& Sun, G. (2018). Squeeze-and-excitation networks. In \emph{Proceedings of the IEEE Conference on Computer Vision and Pattern Recognition} (pp. 7132-7141). IEEE. \url{https://doi.org/10.1109/CVPR.2018.00745}
\item
  Ioffe, S., \& Szegedy, C. (2015). Batch normalization: Accelerating deep network training by reducing internal covariate shift. \emph{In Proceedings of the 32nd International Conference on Machine Learning} (pp. 448-456). PMLR. \url{http://proceedings.mlr.press/v37/ioffe15.html}
\item
  Jain, A. K., and Farrokhnia, F. (1991). Unsupervised texture segmentation using Gabor filters. \emph{Pattern Recognition}, 24(12), 1167-1186.\\
  \url{https://doi.org/10.1016/0031-3203(91)90143-S}
\item
  Ke, G., Meng, Q., Finley, T., Wang, T., Chen, W., Ma, W., Ye, Q., \& Liu, T. Y. (2017). LightGBM: A highly efficient gradient boosting decision tree. In \emph{Advances in Neural Information Processing Systems 30} (pp. 3146-3154). Curran Associates, Inc. \url{https://papers.nips.cc/paper/2017/hash/6449f44a102fde848669bdd9eb6b76fa-Abstract.html}
\item
  Kirillov, A., Mintun, E., Ravi, N., Mao, H., Rolland, C., Gustafson, L., Xiao, T., Whitehead, S., Berg, A. C., Lo, W.-Y., Dollár, P., \& Girshick, R. (2023). \emph{Segment anything}. \textbf{arXiv}. \url{https://arxiv.org/abs/2304.02643}
\item
  Laan Labs. \emph{3D Scanner App} (Version 2.3) {[}Mobile app{]}. Apple App Store. \href{https://apps.apple.com/us/app/3d-scanner-app/id1419913995?platform=mac\&utm_source=chatgpt.com}{3D Scanner App listing}
\item
  Lin, T. Y., Goyal, P., Girshick, R., He, K., \& Dollár, P. (2017). Focal loss for dense object detection. In \emph{Proceedings of the IEEE International Conference on Computer Vision} (pp. 2980-2988). IEEE. \url{https://doi.org/10.1109/ICCV.2017.324}
\item
  Liu, Q., Sun, J., Zhuang, H., Yoon, S.-C., Bowker, B., Yang, Y., Zhang, J., \& Pang, B. (2025). Prediction of raw meat texture and myopathic severity of broiler breast meat with the wooden breast condition by hyperspectral imaging. \emph{British Poultry Science, 66}(5), 606-612. \url{https://doi.org/10.1080/00071668.2025.2471450}
\item
  López, T. A., García, P. A., Tica, L., Pieniazek, F., Schebor, C., \& Sosa, N. (2023). Classification, quality characteristics, sensory perception and texture prediction of wooden breast myopathy in broilers from Argentina. \emph{Journal of Food Science and Technology, 60}(1), 233-242. \url{https://doi.org/10.1007/s13197-022-05608-9}
\item
  Loshchilov, I., \& Hutter, F. (2019). Decoupled weight decay regularization. In \emph{Proceedings of the 7th International Conference on Learning Representations}. OpenReview.net. \url{https://openreview.net/forum?id=Bkg6RiCqY7}
\item
  Lu, T., Liu, Y., Shu, X., Zhang, W., Li, M., Chen, S., Li, P., \& Wang, Z. (2024). Deep-learning classification of chicken woody breast based on bioelectrical impedance characteristics. \emph{Food Science of Animal Products, 2}(3), 9240072. \url{https://doi.org/10.26599/FSAP.2024.9240072}
\item
  Lux Optics Incorporated. (2024). \emph{Halide Mark II} (Version 2.17.5) {[}Mobile app{]}. Apple App Store. \href{https://apps.apple.com/gb/app/halide-mark-iii-pro-camera/id885697368}{Halide App Listing}
\item
  Manjunath, B. S., and Ma, W. Y. (1996). Texture features for browsing and retrieval of image data. \emph{IEEE Transactions on Pattern Analysis and Machine Intelligence}, 18(8), 837-842. \url{https://doi.org/10.1109/34.531803}
\item
  Mudalal, S., Lorenzi, M., Soglia, F., Cavani, C., \& Petracci, M. (2015). Implications of white striping and spaghetti meat abnormalities on quality traits of raw and marinated chicken breast meat. \emph{Animal, 9}(4), 728-734. \url{https://doi.org/10.1017/S175173111400295X}
\item
  Muñoz-Lapeira, M., Lintvedt, T. A., Sanden, K. W., Afseth, N. K., Zomeño, C., Font-I-Furnols, M., Jofré, A., \& Wold, J. P. (2025). Discrimination of normal and wooden breast chicken fillets using NIR, fluorescence and Raman spectroscopy. \emph{Spectrochimica Acta Part A: Molecular and Biomolecular Spectroscopy, 343}, 126463. \url{https://doi.org/10.1016/j.saa.2025.126463}
\item
  Oshina, I., \& Spigulis, J. (2021). Beer-Lambert law for optical tissue diagnostics: current state of the art and the main limitations. \emph{Journal of Biomedical Optics}, 26(10), 100901.\url{https://doi.org/10.1117/1.JBO.26.10.100901}
\item
  Pallerla, C., Feng, Y., Owens, C. M., Bist, R. B., Mahmoudi, S., Sohrabipour, P., Davar, A., \& Wang, D. (2024). Neural network architecture search enabled wide-deep learning (NAS-WD) for spatially heterogenous property awared chicken woody breast classification and hardness regression. \emph{Artificial Intelligence in Agriculture, 14}, 73-85. \url{https://doi.org/10.1016/j.aiia.2024.11.003}
\item
  Pallerla, C., Mahmoudi, S., \& Wang, D. (2026). MyoVision: A mobile research tool and NEATBoost-attention ensemble framework for real time chicken breast myopathy detection. \emph{Proceedings of the IEEE/CVF Conference on Computer Vision and Pattern Recognition Workshops (CVPRW).}
\item
  Pedregosa, F., Varoquaux, G., Gramfort, A., Michel, V., Thirion, B., Grisel, O., Blondel, M., Prettenhofer, P., Weiss, R., Dubourg, V., Vanderplas, J., Passos, A., Cournapeau, D., Brucher, M., Perrot, M., \& Duchesnay, É. (2011). Scikit-learn: Machine learning in Python. \emph{Journal of Machine Learning Research, 12}, 2825-2830. \url{http://jmlr.org/papers/v12/pedregosa11a.html}
\item
  Petracci, M., Mudalal, S., Soglia, F., \& Cavani, C. (2015). Meat quality in fast-growing broiler chickens. \emph{World\textquotesingle s Poultry Science Journal, 71}(2), 363-374. \url{https://doi.org/10.1017/S0043933915000367}
\item
  Qi, C. R., Su, H., Mo, K., \& Guibas, L. J. (2017). PointNet: Deep learning on point sets for 3D classification and segmentation. \emph{In Proceedings of the IEEE Conference on Computer Vision and Pattern Recognition} (pp. 652-660). IEEE. \url{https://doi.org/10.1109/CVPR.2017.16}
\item
  Qi, C. R., Yi, L., Su, H., \& Guibas, L. J. (2017). PointNet++: Deep hierarchical feature learning on point sets in a metric space. In \emph{Advances in Neural Information Processing Systems 30} (pp. 5099-5108). \url{https://proceedings.neurips.cc/paper_files/paper/2017/file/d8bf84be3800d12f74d8b05e9b89836f-Paper.pdf}
\item
  Siddique, A., Herron, C. B., Valenta, J., Garner, L. J., Gupta, A., Sawyer, J. T., \& Morey, A. (2022). Classification and feature extraction using supervised and unsupervised machine learning approach for broiler woody breast myopathy detection. \emph{Foods, 11}(20), 3270. \url{https://doi.org/10.3390/foods11203270}
\item
  Sihvo, H. K., Immonen, K., \& Puolanne, E. (2014). Myodegeneration with fibrosis and regeneration in the pectoralis major muscle of broilers. \emph{Veterinary Pathology, 51}(3), 619-623. \url{https://doi.org/10.1177/0300985813497488}
\item
  Soglia, F., Laghi, L., Canonico, L., Cavani, C., \& Petracci, M. (2016). Functional property issues in broiler breast meat affected by emerging muscle abnormalities. \emph{Food Research International, 89}(Pt 1), 1071-1076. \url{https://doi.org/10.1016/j.foodres.2016.04.042}
\item
  Stanley, K. O., \& Miikkulainen, R. (2002). Evolving neural networks through augmenting topologies. \emph{Evolutionary Computation}, 10(2), 99-127. \url{https://doi.org/10.1162/106365602320169811}
\item
  Tijare, V. V., Yang, F. L., Kuttappan, V. A., Alvarado, C. Z., Coon, C. N., \& Owens, C. M. (2016). Meat quality of broiler breast fillets with white striping and woody breast muscle myopathies. \emph{Poultry Science}, 95(9), 2167-2173. \url{https://doi.org/10.3382/ps/pew129}
\item
  Wold, J. P., Måge, I., Løvland, A., Sanden, K. W., \& Ofstad, R. (2019). Near-infrared spectroscopy detects woody breast syndrome in chicken fillets by the markers protein content and degree of water binding. \emph{Poultry Science, 98}(1), 480-490. \url{https://doi.org/10.3382/ps/pey351}
\item
  Wold, J. P., Veiseth-Kent, E., Høst, V., \& Løvland, A. (2017). \emph{Rapid on-line detection and grading of wooden breast myopathy in chicken fillets by near-infrared spectroscopy}. PLoS ONE, 12(3), e0173384. \url{https://doi.org/10.1371/journal.pone.0173384}
\item
  Wolpert, D. H. (1992). Stacked generalization. \emph{Neural Networks, 5}(2), 241-259. \url{https://doi.org/10.1016/S0893-6080(05)80023-1}
\item
  Xue, W., Zhang, L., Mou, X., \& Bovik, A. C. (2014). Gradient magnitude similarity deviation: A highly efficient perceptual image quality index. \emph{IEEE Transactions on Image Processing}, 23(2), 684-695. \url{https://doi.org/10.1109/TIP.2013.2293423}
\item
  Yang, Y., Wang, W., Zhuang, H., Yoon, S.-C., Bowker, B., Jiang, H., \& Pang, B. (2021). Evaluation of broiler breast fillets with the woody breast condition using expressible fluid measurement combined with deep learning algorithm. \emph{Journal of Food Engineering, 288}, 110133. \url{https://doi.org/10.1016/j.jfoodeng.2020.110133}
\item
  Yoon, S. C., Bowker, B. C., \& Zhuang, H. (2016). \emph{Toward a fusion of image coherence tomography and hyperspectral imaging for poultry meat quality assessment}. \emph{In Proceedings of the IS\&T International Symposium on Electronic Imaging} (Vol. 28, Article 00008). Society for Imaging Science and Technology. \url{https://doi.org/10.2352/ISSN.2470-1173.2016.14.IPMVA-380}
\item
  Yoon, S.-C., Bowker, B. C., Zhuang, H., \& Lawrence, K. C. (2022). Development of imaging system for online detection of chicken meat with wooden breast condition. \emph{Sensors, 22}(3), 1036. \url{https://doi.org/10.3390/s22031036}
\item
  Yu, X., Tang, L., Rao, Y., Huang, T., Zhou, J., \& Lu, J. (2022). Point-BERT: Pre-training 3D point cloud transformers with masked point modeling. In \emph{Proceedings of the IEEE/CVF Conference on Computer Vision and Pattern Recognition (CVPR)}. \url{https://openaccess.thecvf.com/content/CVPR2022/papers/Yu_Point-BERT_Pre-Training_3D_Point_Cloud_Transformers_With_Masked_Point_Modeling_CVPR_2022_paper.pdf}
\item
  Zhou, Z. H. (2012). Ensemble methods: Foundations and algorithms. \emph{Chapman and Hall/CRC}. \url{https://doi.org/10.1201/b12207}
\end{enumerate}

\end{document}